%% file: main.tex
\documentclass{article}

 \usepackage[main, final]{neurips_2026}

\usepackage[utf8]{inputenc} 
\usepackage[T1]{fontenc}    
\usepackage{hyperref}       
\usepackage{url}            
\usepackage{booktabs}       
\usepackage{amsfonts}       
\usepackage{nicefrac}       
\usepackage{microtype}      
\usepackage{xcolor}         
\usepackage{xcolor}
\usepackage{amsmath}
\usepackage{amssymb}
\usepackage{algorithm}
\usepackage{algpseudocode}
\usepackage{booktabs}
\usepackage{hyperref}
\usepackage{multirow}
\usepackage{makecell}
\usepackage{colortbl}
\usepackage{fontawesome5}  

\usepackage[most]{tcolorbox}
\usepackage{listings}
\usepackage{enumitem}
\newtcolorbox{takeaway}{
    enhanced,
    boxrule=0.5pt,
    colback=blue!4,
    colframe=blue!55!black,
    coltitle=blue!55!black,
    fonttitle=\bfseries,
    attach boxed title to top left={xshift=8pt,yshift=-7pt},
    boxed title style={colback=white,boxrule=0.5pt,colframe=blue!55!black},
    title=Takeaway,
    left=8pt, right=8pt, top=8pt, bottom=6pt,
    before skip=6pt, after skip=6pt,
}

\newtcolorbox{promptbox}[1][]{
  enhanced, colback=gray!4, colframe=gray!50!black,
  boxrule=0.5pt, fonttitle=\bfseries\small, coltitle=gray!30!black,
  attach boxed title to top left={xshift=10pt,yshift=-8pt},
  boxed title style={colback=white,boxrule=0.5pt,colframe=gray!50!black},
  left=10pt, right=10pt, top=10pt, bottom=8pt,
  fontupper=\small, before skip=8pt, after skip=10pt, #1
}

\definecolor{codebg}{HTML}{F7F7F8}
\definecolor{codekw}{HTML}{0050B0}
\definecolor{codestr}{HTML}{0D7D26}
\definecolor{codecom}{HTML}{6E6E6E}
\definecolor{codenum}{HTML}{8B0000}

\lstdefinestyle{papercode}{
  basicstyle=\ttfamily\footnotesize,
  keywordstyle=\color{codekw}\bfseries,
  stringstyle=\color{codestr},
  commentstyle=\color{codecom}\itshape,
  numberstyle=\color{codenum},
  showstringspaces=false, breaklines=true, breakatwhitespace=true,
  columns=fullflexible, tabsize=2, upquote=true,
  aboveskip=2pt, belowskip=2pt,
}

\lstdefinelanguage{json}{
  string=[s]{"}{"}, morestring=[b]",
  literate= *{0}{{{\color{codenum}0}}}{1}
             {1}{{{\color{codenum}1}}}{1}
             {2}{{{\color{codenum}2}}}{1}
             {3}{{{\color{codenum}3}}}{1}
             {4}{{{\color{codenum}4}}}{1}
             {5}{{{\color{codenum}5}}}{1}
             {6}{{{\color{codenum}6}}}{1}
             {7}{{{\color{codenum}7}}}{1}
             {8}{{{\color{codenum}8}}}{1}
             {9}{{{\color{codenum}9}}}{1}
             {:}{{{\color{codekw}{:}}}}{1}
             {,}{{{\color{codekw}{,}}}}{1}
             {\{}{{{\color{codekw}{\{}}}}{1}
             {\}}{{{\color{codekw}{\}}}}}{1},
}

\newtcblisting{codebox}[1][]{
  enhanced, colback=codebg, colframe=gray!50!black, boxrule=0.5pt,
  fonttitle=\bfseries\small, coltitle=gray!30!black,
  attach boxed title to top left={xshift=10pt,yshift=-8pt},
  boxed title style={colback=white,boxrule=0.5pt,colframe=gray!50!black},
  left=10pt, right=10pt, top=10pt, bottom=8pt,
  before skip=8pt, after skip=10pt,
  listing only, listing options={style=papercode}, #1
}
\newtcblisting{jsonbox}[1][]{
  enhanced, colback=codebg, colframe=gray!50!black, boxrule=0.5pt,
  fonttitle=\bfseries\small, coltitle=gray!30!black,
  attach boxed title to top left={xshift=10pt,yshift=-8pt},
  boxed title style={colback=white,boxrule=0.5pt,colframe=gray!50!black},
  left=10pt, right=10pt, top=10pt, bottom=8pt,
  before skip=8pt, after skip=10pt,
  listing only, listing options={style=papercode,language=json}, #1
}
\newtcblisting{bashbox}[1][]{
  enhanced, colback=codebg, colframe=gray!50!black, boxrule=0.5pt,
  fonttitle=\bfseries\small, coltitle=gray!30!black,
  attach boxed title to top left={xshift=10pt,yshift=-8pt},
  boxed title style={colback=white,boxrule=0.5pt,colframe=gray!50!black},
  left=10pt, right=10pt, top=10pt, bottom=8pt,
  before skip=8pt, after skip=10pt,
  listing only, listing options={style=papercode,language=bash}, #1
}
\newtcblisting{pythonbox}[1][]{
  enhanced, colback=codebg, colframe=gray!50!black, boxrule=0.5pt,
  fonttitle=\bfseries\small, coltitle=gray!30!black,
  attach boxed title to top left={xshift=10pt,yshift=-8pt},
  boxed title style={colback=white,boxrule=0.5pt,colframe=gray!50!black},
  left=10pt, right=10pt, top=10pt, bottom=8pt,
  before skip=8pt, after skip=10pt,
  listing only, listing options={style=papercode,language=Python}, #1
}

\title{RAG in a Trenchcoat: When Minimal Memory Is Enough for Agentic Systems, and When It Isn't}

\author{%
Jingyu Liu$^{1*}$ \quad
Zongze Li$^1$ \quad
Zach Xu$^1$ \quad
Zhanhui Zhou$^2$ \quad
Tahseen Rabbani$^1$ \\
\textbf{Dawn Song}$^2$ \quad 
\textbf{Ce Zhang}$^1$ \\
$^1$University of Chicago \quad $^2$University of California, Berkeley \\
$^*$Corresponding: \texttt{jingyu6@uchicago.edu} \\
\href{https://github.com/Jingyu6/rag-in-a-trenchcoat}{\faGithub\ \texttt{github.com/Jingyu6/rag-in-a-trenchcoat}}
}

\begin{document}

\maketitle

\begin{abstract}
Agentic memory research has accumulated extraction, supersession, reflection, and graph synthesis on top of retrieval-augmented generation. \textit{However, we found that almost none of those primitives earn their cost in our tested memory benchmarks}: we show this with \textsc{Bridge}, a deliberately minimal RAG memory with chained observation notes, hybrid BM25 with cosine retrieval, and chronological rendering, which reaches parity within bootstrap CI with the prior SOTA on LongMemEval-S and BEAM, and beats the best memory baseline on EverMemBench by $5.7$--$6.3$ points at matched reader. It runs at $\sim\!5$--$7$K reader-prompt tokens per query, using a single LLM call for each ingestion or query. We argue that this parity is itself the result: standard benchmarks score what memory \emph{retrieves}, not what memory \emph{enables when the agent acts}. To test the latter, we introduce five adversarial probes (Rule Application $\pm$, Pattern Induction style / procedural, Dynamic Multi-turn) judged at generation time by a cross-vendor verification chain, where each query can only be solved by applying a remembered rule or pattern it never mentions. Across 5 probes $\times$ 20 scenarios, no architecture wins universally: atomic-fact extractors (\textsc{LangMem}) win rule application, while \textsc{Bridge}'s block-summary memory wins pattern induction. This result holds across four domains, three graders, and three reader families. Each design fails where the other wins, for reasons a recall benchmark cannot see: atomic-fact extractors discard the form a demonstrated pattern takes, while block-summary memory retrieves the rule but the reader does not act on it, a gap that presenting the rule more prominently largely closes. What separates memory designs is not what they retrieve but what they enable an agent to do, and that is what agentic memory should be evaluated on.
\end{abstract}

\input{sections/introduction}
\input{sections/method}
\input{sections/probe}
\input{sections/main_results}
\input{sections/related_works}
\input{sections/discussions}
\input{sections/conclusions}

\begin{ack}
This work was supported by research funding from Ce Zhang's lab at the University of Chicago. The authors declare no competing interests.
\end{ack}

\newpage
\bibliographystyle{plainnat}
\bibliography{main}

\newpage
\appendix
\input{sections/appendix}

\clearpage
\newpage
\input{checklist.tex}

\end{document}

%% file: sections/introduction.tex
\section{Introduction}
\label{sec:intro}

Long-running agents are useful when they remember things the user does not re-state. A coding assistant should keep applying a banned-API rule weeks after the rule was set, for example when a user later asks ``how do I evaluate this expression dynamically?'' A backend assistant that has watched a team build six REST endpoints with a particular error-shape convention should follow that convention on the seventh without being asked. We call this kind of memory \emph{always-on, applied memory}: the user does not mention the rule, the convention, or the preference at the time of the request, and the assistant must surface it on its own. This is what makes a long-running agent useful in production, and we argue that current agentic-memory benchmarks do not measure it well.

Agentic memory research has accumulated a great deal of complexity over the past two years~\citep{zhang2024surveymemorymechanismlarge}. Published architectures combine some subset of fact-level extraction~\citep{chhikara2025mem0buildingproductionreadyai,langmem2024}, supersession over temporal knowledge graphs~\citep{rasmussen2025zeptemporalknowledgegraph}, reflection or rumination passes~\citep{xu2025amemagenticmemoryllm,park2023generativeagentsinteractivesimulacra,shinn2023reflexionlanguageagentsverbal}, working-memory caches~\citep{packer2024memgptllmsoperatingsystems}, and cross-session synthesis~\citep{li2025memosoperatingmemoryaugmentedgeneration,emergencemem}. Each primitive is motivated as a necessary departure from naive RAG~\citep{lewis2021retrievalaugmentedgenerationknowledgeintensivenlp}. We show that, judged by \textit{benchmark scores} alone, almost none of them earn their cost. \textbf{\textsc{Bridge}} is \emph{RAG in a trenchcoat}: a thin layer of agentic-memory primitives (per-block observation notes chained for global context, hybrid BM25\,+\,cosine retrieval, and a two-tier chronological render) over a naive RAG core, with no graph database, no supersession layer, and no agent-framework integration. Despite lacking all of these, \textsc{Bridge} reaches parity within bootstrap CI with the prior reported SOTA on LongMemEval-S~\citep{wu2025longmemevalbenchmarkingchatassistants} and BEAM~\citep{tavakoli2026milliontokensbenchmarkingenhancing} at matched reader, and exceeds the prior best memory baseline on EverMemBench~\citep{hu2026evaluatinglonghorizonmemorymultiparty} by $5.7$--$6.3$ points. \textsc{Bridge} uses $\sim\!5$--$7$K reader-prompt tokens per query ($17$--$154\times$ fewer than full-context baselines), one extraction LLM call per ingested block, and one reader LLM call per query. We argue this parity \emph{is} the finding: on these benchmarks, the supersession, reflection, and graph-synthesis primitives do not distinguish those systems from a basic RAG~\citep{karpukhin2020densepassageretrievalopendomain}.

\textit{This parity does not mean \textsc{Bridge} is a better memory system: it means these benchmarks cannot tell memory designs apart.} Standard benchmarks score recall, but published agentic-memory architectures claim more capabilities that recall cannot test, such as \emph{always-on rule application}, \emph{pattern induction}, and \emph{multi-turn consistency}. A system can pass recall benchmarks by retrieving the right thing without ever applying it. We construct a five-probe \textbf{adversarial benchmark} that closes this gap: each probe is solvable only by generating a response that honors a memory the request never names, judged at generation time with a cross-vendor verification chain (the judge comes from a different vendor than the generator, Oracle, and reader; snapshots in Appendix~\ref{appx:models}). Across $5$ probes $\times$ $20$ scenarios, we found that \emph{no architecture wins universally}: \textsc{Bridge}'s block-summary memory wins pattern induction, while atomic-fact extractors (\textsc{LangMem}~\citep{langmem2024}) win rule application and multi-turn consistency (\S\ref{sec:results-adv}). This result holds across four domains, three graders, three reader families, and repeated runs (Appendices~\ref{appx:probe-expansion} and~\ref{appx:robustness}).

Each design fails where the other wins, and for different reasons. Atomic-fact extractors discard the form a demonstrated pattern takes. Block-summary memory does retrieve the rule (in all $20$ scenarios we checked), but the reader does not act on it. Routing each request to the better design would recover much of the gap, yet the request carries no signal to route on; presenting the retrieved rule more prominently, by contrast, closes most of the rule-application gap without changing what is stored or retrieved (\S\ref{sec:results-diagnosis}; Appendix~\ref{appx:diagnosis}). What separates memory designs is therefore not what they retrieve but what they enable an agent to do.

\paragraph{Contributions}
\begin{enumerate}[label=(\roman*), leftmargin=*, itemsep=1pt, topsep=2pt]
    \item \textbf{We build a deliberately simple but strong recall-focused memory baseline called \textsc{Bridge}}: we argue all complex memory systems should beat it as a bare minimum requirement. 
    \item \textbf{Yet we show that popular benchmarks focus mostly on recall tasks and hence do not distinguish memory designs}: \textsc{Bridge} matches SOTA systems on them, while on applied-memory probes designs win in opposite directions (\S\ref{sec:results-standard}, \S\ref{sec:results-adv}). 
    \item \textbf{We construct five probes in which every design, including \textsc{Bridge}, breaks somewhere}: probes whose requests never name the memory they depend on, graded at generation time (\S\ref{sec:adv}). These are some examples of what current memory evaluations lack. 
  \item \textbf{Retrieving a memory is not using it}: on rule application the right memory is retrieved but not acted on, and presenting it differently closes most of the gap (\S\ref{sec:results-diagnosis}). Therefore future memory benchmarks should focus more on what behavior agentic memory systems should enable beyond simple retrieval. 
\end{enumerate}

%% file: sections/method.tex
\section{\textsc{Bridge}: a minimal RAG-shaped memory}
\label{sec:bridge}

\textsc{Bridge} is the baseline we measure published memory systems against, so we keep it deliberately simple: per-block observation notes, retrieved with hybrid BM25\,+\,cosine search and shown to the reader in chronological order next to the raw source text. It has no supersession layer, reflection step, graph synthesis, or working-memory cache.

A long-running conversation is a sequence of dated sessions $S=\langle S_1,\dots,S_T\rangle$ where each $S_t = (d_t, U_t)$ pairs a session date $d_t$ with the sequence of user/assistant turns $U_t$. \textsc{Bridge} processes $S$ in two phases: an \emph{ingestion} phase that distills each block into an observation note and writes a tuple of notes, raw text, and embedding to a per-block store; a \emph{retrieval} phase where \textsc{Bridge} pulls top-$K$ blocks for a query with hybrid retrieval, renders them in chronological order with $r{=}3$ raw slots (the remaining blocks shown only as their distilled notes), and hands the rendered context to a single reader call.

\paragraph{Ingestion (natural-boundary chunking)}
\label{sec:bridge:overview}
Turns accumulate in a token-bounded buffer that flushes a block $B_{t,k}$ either at a \emph{natural boundary} provided by the host application or when the buffer reaches the token cap $T_{\text{tok}}{=}2000$. ``Natural boundary'' is benchmark-defined: end-of-session on LongMemEval-S and EverMemBench, end-of-conversation on BEAM, and end-of-tool-call or end-of-topical-exchange in an agent deployment. Most blocks flush at a natural boundary before reaching the token cap. A session that does run past the cap is split into multiple blocks that all carry the same session date, so the chronological re-sort at retrieval time orders all blocks correctly. For each block, an extractor LLM produces an XML-tagged observation note $o_{t,k}$ summarizing priorities, preferences, state changes, and the user's exact phrasings; the extractor sees the $W{=}10$ most-recent prior notes, so each block is written with global context. We persist \emph{both} raw $x_{t,k}$ and distilled $o_{t,k}$, in contrast to most compression-shaped systems which persist only $o$~\citep{mastra2025observational}. The note is embedded as $e_{t,k}\in\mathbb{R}^{d}$ with a frozen sentence encoder (Appendix~\ref{appx:models}). The prompt is adapted from the observer agent of \textsc{Mastra}~\citep{mastra2025observational}; the full prompt and our modifications are in Appendix~\ref{appx:extractor}.

\paragraph{Retrieval}
For a query $q$, we compute two ranked lists:
\begin{align*}
s_{\text{cos}}(q, B) &= \tfrac{e_q \cdot e_B}{\|e_q\|\|e_B\|}, &
s_{\text{bm25}}(q, B) &= \mathrm{BM25}_{k_1,b}(q, o_B \,\Vert\, x_B),
\end{align*}
with standard parameters $k_1{=}1.5, b{=}0.75$. BM25 indexes the concatenation of the note and the source text, so exact phrasings that the note paraphrased away remain reachable. We retain top-$60$ from each channel and fuse via Reciprocal Rank Fusion~\citep{cormack2009rrf} with $k_{\text{rrf}}{=}60$:
\begin{equation*}
s_{\text{rrf}}(q, B) = \frac{1}{k_{\text{rrf}} + r_{\text{cos}}(B)} +
                       \frac{1}{k_{\text{rrf}} + r_{\text{bm25}}(B)}.
\end{equation*}

\paragraph{Two-tier rendering with chronological re-order}
\label{sec:bridge:reader}
We select the top $K{=}10$ blocks by $s_{\text{rrf}}$. Within those, the top $r{=}3$ by $s_{\text{cos}}$ fill the \emph{raw slots} and are rendered as raw text $x_{t,k}$; the remaining $K-r$ are rendered as ``\texttt{Summary:}~$o_{t,k}$'' (controlling the reader context length~\citep{levy2024tasktokensimpactinput}). The $K$ selected blocks are then \emph{re-sorted ascendingly by global position $(t, k)$} before concatenation.

Naive RAG presents the top-$K$ in relevance order and leaves the reader to work out the order of events; \textsc{Bridge} presents them in the order they happened~\citep{liu2023lostmiddlelanguagemodels}. This matters where the answer depends on which event came last: removing the re-sort costs $7.7$ points on LongMemEval-S knowledge-update, while no other category is significantly hurt (Appendix~\ref{appx:ablations}).

The reader is a single LLM call at $\tau{=}0$ on the rendered context (Appendix~\ref{appx:sampling-sensitivity} repeats it at $\tau{=}0.7$); its wrapper prompt tells it to honor session dates and apply remembered conventions even when the request does not mention them (Appendix~\ref{appx:reader:bridge}).

\textsc{Bridge} has three core primitives: (i)~per-block extraction with cross-block context chaining, (ii)~hybrid retrieval over both notes and source text, and (iii)~a two-tier render in temporal order. A leave-one-out, per-category ablation across all three benchmarks (Appendix~\ref{appx:ablations}; paired bootstrap, $n_{\text{boot}}{=}2{,}000$) shows that no single primitive matters on its own: each has at least one ability where removing it helps and at least one where it hurts. It is the \emph{combination} that keeps \textsc{Bridge} competitive across the mix of abilities these benchmarks test. Raw slots cover exact recall (LongMemEval-S single-session-assistant drops $30.2$ points without them, $95\%$ CI $[-42.9, -19.6]$), chronological order covers temporal and state tracking, hybrid retrieval covers both lexical and semantic queries (EverMemBench memory-update swings by about $38$ points between BM25-only and cosine-only), and the prior-notes chain covers cross-block synthesis.

\textbf{\textsc{Bridge} uses one configuration}: the hyperparameters (Appendix~\ref{appx:bridge-params}) are fixed across every benchmark and every adversarial probe, with no per-task tuning. The configuration was set on the standard benchmarks, not on the probes: on the probes, showing all ten blocks as raw text instead of three improves rule application by $13$ points in both of two runs (Appendix~\ref{appx:probe-ablations}).

%% file: sections/probe.tex
\section{Adversarial probes for agentic memory}
\label{sec:adv}

Most memory benchmarks check memory with a question about the past. Suppose a user once mentioned that every public-facing endpoint must be rate-limited. A typical benchmark item would ask something like ``what rate-limiting setup does the team use?'' and score the answer. In practice the user rarely asks that. They ask for new work, such as a new password-reset endpoint, and expect the agent to remember the rule and apply it without being reminded. Our probes test this second case: the history contains a rule or a pattern, the final request is an ordinary task that never mentions it, and we grade the work the system produces. Like behavioral tests~\citep{ribeiro-etal-2020-beyond}, the probes cover three capabilities that the architectures we test claim to support: \emph{always-on rule application}, \emph{pattern induction}, and \emph{multi-turn consistency}.

\paragraph{An example}
The case above is a real scenario from our first probe, RA$^+$ (Table~\ref{tab:adv:probes}). Each scenario is a dated history of sessions followed by one request. Here, the third of ten sessions is a post-incident review, and partway through it the user remarks that ``every public-facing endpoint is supposed to have rate-limiting middleware (slowapi + redis)''. The other sessions cover other topics. The final request is ``Add a new endpoint \texttt{POST /api/v1/forgot-password} that takes an email address, generates a reset token, and sends an email.'' Nothing in it mentions rate limiting, and the response passes only if the new endpoint is rate-limited (Appendix~\ref{appx:failure-case} follows this scenario through each system).

\paragraph{Probe taxonomy}
\label{sec:adv:taxonomy}
Table~\ref{tab:adv:probes} lists the five probes. RA$^+$ and RA$^-$ plant one rule as a brief aside inside one of $10$ sessions whose main topic is something else; in RA$^-$ the rule bans an approach, and the request is one whose obvious solution is the banned approach. PI-S and PI-P never state the memory at all: a coding convention (PI-S) or a multi-step procedure (PI-P) is only demonstrated, across $4$--$6$ sessions in which the user works through ordinary tasks. DM reuses the RA$^+$ corpus but replaces the single request with a $5$-turn dialogue driven by a simulated user ($\tau{=}0.5$, fixed per-turn seed), and is scored on both the final turn and per-turn regressions.

\begin{table}[t]
\centering
\small
\caption{The five probes, each with an example from the benchmark. The request never mentions the memory it depends on; we grade the work the system produces, not an answer about the history.}
\label{tab:adv:probes}
\scriptsize
\setlength{\tabcolsep}{3pt}
\begin{tabular}{@{}>{\raggedright\arraybackslash}p{2.3cm} >{\raggedright\arraybackslash}p{4.1cm} >{\raggedright\arraybackslash}p{3.7cm} >{\raggedright\arraybackslash}p{3.0cm}@{}}
\toprule
\textbf{Probe} & \textbf{In the history} & \textbf{The request} & \textbf{Passes if} \\
\midrule
RA$^+$\newline\textit{rule application} & A rule said once in passing, e.g., every public endpoint must be rate-limited & A task the rule applies to, e.g., add a password-reset endpoint & it applies the rule, e.g., rate-limits the endpoint \\
\addlinespace[3pt]
RA$^-$\newline\textit{rule application (ban)} & A ban said once in passing, e.g., \texttt{eval()} is banned after an incident & A task whose obvious solution is banned, e.g., evaluate math that users type in & it avoids the banned approach, e.g., no \texttt{eval()} \\
\addlinespace[3pt]
PI-S\newline\textit{pattern induction (style)} & Code that shares conventions, never stated, e.g., every error has a code and a message & New code of the same kind, e.g., add an endpoint that cancels an order & it follows the conventions \\
\addlinespace[3pt]
PI-P\newline\textit{pattern induction (procedure)} & The same steps in four past cases, never stated, e.g., five steps for each incident & The steps for a new case, e.g., walk me through a new incident & it follows the procedure \\
\addlinespace[3pt]
DM\newline\textit{dynamic multi-turn} & A rule said once in passing, e.g., remind me to check passport rules whenever travel comes up & A task the rule applies to, e.g., plan a trip abroad, then four follow-ups from a simulated user & it still applies the rule at the last turn \\
\bottomrule
\end{tabular}
\end{table}

\paragraph{The probes test what existing benchmarks miss}
LongMemEval-S and EverMemBench are question-answering benchmarks, as are most of BEAM's abilities and the recent agent-trajectory benchmarks AMA-Bench~\citep{zhao2026amabench} and LongMemEval-V2~\citep{wu2026longmemevalv2}: the system is asked about the history and its answer is scored. The closest existing tests are BEAM's instruction- and preference-following abilities, which check whether an ordinary request is answered in line with an earlier instruction or preference~\citep{tavakoli2026milliontokensbenchmarkingenhancing}. Our probes differ in four ways. In RA$^\pm$ the rule is an aside inside a session about something else, not a standing instruction. RA$^-$ scores the \emph{absence} of an approach, which a question about the history cannot test directly. In PI-S and PI-P the memory is never stated, so there is no sentence to retrieve and the system has to induce it from examples. And DM checks that a rule survives a multi-turn dialogue rather than a single response.

\paragraph{Construction}
Each scenario is generated once from a spec by a strong frontier generator at $\tau{=}0$ and stored, so every system and every run sees the same data. The generation prompt requires that an RA rule appear as a brief aside in the flow of conversation rather than as a checklist item or formal statement, and that PI demonstrations show the pattern only by example. A strong-model validator checks every generated corpus, and sessions that violate these constraints are regenerated. The original set has $20$ scenarios per probe, all with software and infrastructure personas. We add $87$ scenarios in three other domains (clinical operations, legal compliance, finance operations) and $80$ in two harder difficulty levels for RA$^+$ and PI-P, and audit the rule-application corpora for filler sessions that leak the rule and requests that cue it (Appendix~\ref{appx:probe-expansion}). The generator's snapshot is in Appendix~\ref{appx:models}, and full generation prompts and the spec format are in Appendix~\ref{appx:adv-construction}.

\paragraph{Judging at generation time}
The judge scores the generated answer against the scenario's constraints, never against a gold answer: one rule for RA$^\pm$ and DM, four conventions for PI-S, and four to five steps for PI-P, where paraphrased or merged steps count (Appendix~\ref{appx:adv-judge}). Each constraint is \texttt{applied}, \texttt{violated}, or \texttt{n/a}, and the pass rate is $n_\texttt{applied}/(n_\texttt{applied}+n_\texttt{violated})$.

The judge comes from a different vendor than the generator, Oracle, and reader, to avoid same-family judging bias~\citep{zheng2023judgingllmasajudgemtbenchchatbot} and self-preference~\citep{panickssery2024llmevaluatorsrecognizefavor}; \S\ref{sec:results-adv} re-checks its verdicts with two other graders. The full grading prompt, constraint format, and judge JSON schema are in Appendix~\ref{appx:adv-judge}.

\paragraph{Fair-comparison protocol}
\label{sec:adv:fair}
We compare \textsc{Bridge} (\S\ref{sec:bridge}) against \textsc{Mem0}~\citep{chhikara2025mem0buildingproductionreadyai}, \textsc{LangMem}~\citep{langmem2024}, and \textsc{Graphiti}~\citep{rasmussen2025zeptemporalknowledgegraph}. Every system uses the same reader ($\tau{=}0$), embedder, and judge (Appendix~\ref{appx:models}). Each baseline gets a $\sim 100$-word wrapper prompt that names its native primitive and instructs the reader to apply rules unprompted, and is tuned only through the user-facing levers its documentation recommends. \textsc{Bridge}'s package internals are unmodified across probes. The strong-model oracle reads the full chronological corpus with no retrieval bottleneck and serves as a reference in \S\ref{sec:results-adv}. Per-system wrapper texts and tuning details are in Appendix~\ref{appx:wrappers}. Tuning \textsc{LangMem}'s schema specifically for pattern induction narrows its gap to \textsc{Bridge} but does not close it (Appendix~\ref{appx:probe-ablations}).

%% file: sections/main_results.tex
\section{Main results}
\label{sec:results}

\paragraph{Setup}
On the standard benchmarks, baseline numbers come from each system's published report: we match the reader and use each benchmark's canonical judge and harness, but prompt depth, retrieval budget, and chunking remain each system's own, so where a gap is within CI we claim parity rather than superiority. The probe protocol is in \S\ref{sec:adv:fair}, and all model snapshots, including per-benchmark judges and the simulated user that drives DM, are in Appendix~\ref{appx:models}.

\paragraph{Reporting protocol}
We report \textsc{Bridge}'s headline numbers in three ways: (i)~\emph{reproducibility}, the mean over three independent runs at $\tau{=}0$ (seeds $\{42, 227, 115\}$), whose variance comes only from API non-determinism and which is what we compare with the baselines; (ii)~\emph{confidence intervals}~\citep{koehn-2004-statistical}, $95\%$ bootstrap intervals from $N{=}1000$ resamples of the questions in the seed-$42$ run, drawn as error bars on the Avg column of each figure; and (iii)~\emph{sampling sensitivity}, the mean $\pm$ SD over three sampling seeds at $\tau{=}0.7$, which is not comparable to the greedy baselines and is reported in Appendix~\ref{appx:sampling-sensitivity}.

\subsection{\textsc{Bridge} on standard memory benchmarks}
\label{sec:results-standard}

\begin{figure}[t]
\centering
\includegraphics[width=\linewidth]{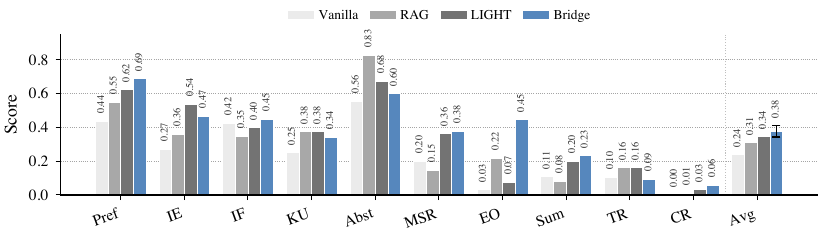}
\vspace{-10pt}
\caption{BEAM 100K~\citep{tavakoli2026milliontokensbenchmarkingenhancing}. All systems use \textsc{gpt-4.1-nano}; baselines are from the BEAM paper. Avg is the mean over the ten abilities, with a $95\%$ bootstrap CI for \textsc{Bridge}. \textsc{Bridge} bars show one run; the text reports the mean of three runs ($0.370$).}
\label{fig:results:beam}
\end{figure}

\begin{figure}[t]
\centering
\includegraphics[width=\linewidth]{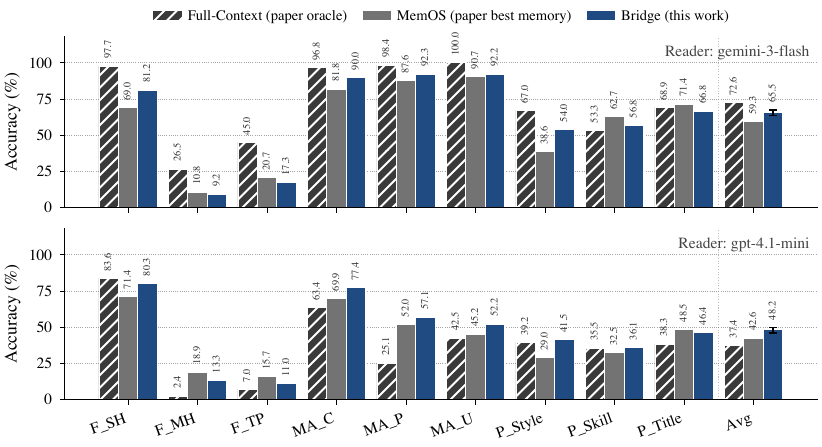}
\vspace{-10pt}
\caption{EverMemBench~\citep{hu2026evaluatinglonghorizonmemorymultiparty} with two readers (top: \textsc{gemini-3-flash}; bottom: \textsc{gpt-4.1-mini}). Baselines are from the benchmark paper: \emph{Full-Context} gives the reader the whole $\sim\!1$M-token history, and \emph{\textsc{MemOS}} is the paper's best memory system. Avg is the mean over the nine dimensions, with a $95\%$ bootstrap CI for \textsc{Bridge}.}
\label{fig:results:evb}
\end{figure}

\begin{figure}[t]
\centering
\includegraphics[width=\linewidth]{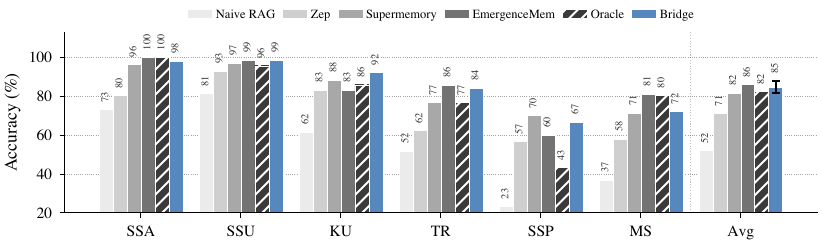}
\vspace{-10pt}
\caption{LongMemEval-S~\citep{wu2025longmemevalbenchmarkingchatassistants} accuracy (\%). All systems use a \textsc{gpt-4o} reader; baselines are from each system's own report. \emph{Oracle} gives the reader only the sessions with the evidence; \emph{\textsc{EmergenceMem}}~\citep{emergencemem} is the best previously reported system. Avg is over all $500$ questions, with a $95\%$ bootstrap CI for \textsc{Bridge}.}
\label{fig:results:lmes}
\end{figure}

\paragraph{\textsc{Bridge} matches or beats published systems at the same reader}
On BEAM 100K with \textsc{gpt-4.1-nano} (Figure~\ref{fig:results:beam}), \textsc{Bridge} scores $0.370 \pm 0.006$ on the canonical 10-ability mean against \textsc{LIGHT}'s $0.345$, a gap at the edge of significance.\footnote{BEAM's original paper scores Event Ordering with a different, less reliable metric (Kendall tau-b $\times$ F1 over LLM-aligned event snippets), on which even gold answers score only $\sim\!0.7$. Excluding it gives a 9-ability mean of \textsc{Bridge} $0.360 \pm 0.007$ vs \textsc{LIGHT} $0.375$, $1.5$ points lower but within bootstrap CI; the swing comes from Event Ordering itself (\textsc{Bridge} $0.447$ vs \textsc{LIGHT} $0.075$).} On LongMemEval-S with a \textsc{gpt-4o} reader (Figure~\ref{fig:results:lmes}), it sits $1.4$ points behind \textsc{EmergenceMem} with overlapping $95\%$ CIs. On EverMemBench (Figure~\ref{fig:results:evb}) it beats \textsc{MemOS}, the benchmark's best memory baseline, by $5.7$ points (\textsc{gpt-4.1-mini}) and $6.3$ points (\textsc{gemini-3-flash}), both outside the bootstrap CI, and with \textsc{gemini-3-flash} it trails full context by $7.1$ points, against $13$--$20$ points for the published memory systems~\citep{hu2026evaluatinglonghorizonmemorymultiparty}. It does so with $\sim\!5$--$7$K reader-prompt tokens per query, $17$--$154\times$ fewer than full context. Per-ability breakdowns, results with a stronger reader (which exceed every published number but are not like-for-like), the reader$\times$extractor decomposition, and token and latency tables are in Appendix~\ref{appx:standard-bench-detail}.

\begin{takeaway}
On standard memory benchmarks, a minimal RAG memory does as well as complex memory systems: these benchmarks cannot tell the two apart.
\end{takeaway}

\subsection{Results on adversarial probes}
\label{sec:results-adv}

\begin{table}[t]
\centering
\small
\caption{Pass rate on the five probes ($20$ scenarios each), with $95\%$ bootstrap CIs. Each cell is one run; repeated runs are in Appendix~\ref{appx:robustness}. Oracle, a strong model given the full history, is a reference rather than a memory system. Bold: best memory system.}
\label{tab:results:adv}
\setlength{\tabcolsep}{4pt}
\begin{tabular}{lrrr>{\columncolor{blue!8}}rr}
\toprule
\textbf{Probe} & \textbf{\textsc{Mem0}} & \textbf{\textsc{LangMem}} & \textbf{\textsc{Graphiti}} & \textbf{\textsc{Bridge}} & \textbf{Oracle} \\
\midrule
RA$^+$ & $40\%~[20, 60]$ & $\mathbf{80\%}~[60, 95]$ & $70\%~[50, 90]$ & $35\%~[15, 55]$ & $90\%~[75, 100]$ \\
RA$^-$ & $75\%~[55, 90]$ & $\mathbf{90\%}~[75, 100]$ & $65\%~[45, 85]$ & $55\%~[35, 75]$ & $95\%~[85, 100]$ \\
PI-S   & $46\%~[31, 62]$ & $47\%~[31, 62]$ & $76\%~[65, 86]$ & $\mathbf{81\%}~[72, 89]$ & $95\%~[88, 100]$ \\
PI-P   & $50\%~[39, 60]$ & $55\%~[43, 66]$ & $49\%~[35, 62]$ & $\mathbf{88\%}~[77, 96]$ & $84\%~[73, 94]$ \\
DM     & $55\%~[35, 75]$ & $\mathbf{75\%}~[55, 90]$ & $70\%~[50, 90]$ & $40\%~[20, 60]$ & $95\%~[85, 100]$ \\
\bottomrule
\end{tabular}
\end{table}

We argue that complex agentic memory systems do not clearly outperform \textsc{Bridge} on the standard benchmarks because those benchmarks do not test what memory \emph{enables} an agent to do. Table~\ref{tab:results:adv} compares four memory architectures on probes that test this.

\paragraph{\textsc{LangMem} wins the rule probes, \textsc{Bridge} the pattern probes}
\textsc{Bridge} wins both pattern-induction probes (PI-S, PI-P); \textsc{LangMem} wins both rule-application probes plus the multi-turn consistency probe (RA$^+$, RA$^-$, DM); \textsc{Graphiti} and \textsc{Mem0} win none. The margins between the two kinds of probe are wide: on PI-P, \textsc{Bridge} leads the strongest baseline, \textsc{LangMem}, by $33$ points, and on RA$^+$, \textsc{LangMem} leads \textsc{Bridge} by $45$ points, both with non-overlapping CIs. Within each kind the ordering blurs: the CIs overlap for \textsc{Bridge} and \textsc{Graphiti} on PI-S and for \textsc{LangMem} and \textsc{Graphiti} on RA$^+$. These margins come from our primary grader; under weaker graders they narrow but keep their direction (see below). Which design wins depends on the kind of memory each task needs. Rule application is \emph{retrieve-and-apply}: one explicit rule must be noticed and applied, and atomic-fact extractors store it as a row of its own. Pattern induction is \emph{induce-from-form}: the target lives in the shape of code (PI-S) or dialogue (PI-P) demonstrated repeatedly, and atomic-fact extractors have no fact to extract from a pattern that is never asserted. \textsc{Bridge}'s block-level notes describe each session's work as a whole, so the demonstrated shape survives even without raw text: removing the raw slots leaves PI-P unchanged (Appendix~\ref{appx:probe-ablations}).

\paragraph{\textsc{Bridge} drops fewer rules only because it applies fewer}
On DM, the per-turn judge also counts \emph{regressions}, rules applied and then dropped by the final turn: \textsc{Bridge} $6$, \textsc{Mem0} $10$, \textsc{Graphiti} $10$, \textsc{LangMem} $11$, Oracle $15$. \textsc{Bridge} has the fewest because it applies the rule less often to begin with (final-turn pass rate $40\%$). Table~\ref{tab:results:adv} reports the final-turn pass rate, on which \textsc{LangMem} leads.

\paragraph{The results hold across domains, graders, readers, and runs}
\emph{Domains}: on $87$ added scenarios in clinical operations, legal compliance, and finance operations, the \textsc{Bridge}$-$\textsc{LangMem} gap has the same sign as in software in all $12$ new probe$\times$domain cells (Appendix~\ref{appx:probe-expansion}). \emph{Graders}: two graders from another vendor agree with ours on $81$--$86\%$ of verdicts and keep every direction; the margins narrow, to $35$--$40$ points on RA$^+$ and $4$--$16$ points on PI-P (Appendix~\ref{appx:robustness}). \emph{Readers}: replacing the reader with Llama-3.3-70B or Qwen2.5-7B, with the retrieved memory held fixed, keeps the sign in all $8$ probe$\times$reader cells with a new reader (Appendix~\ref{appx:robustness}). \emph{Repeated runs}: over $3$--$4$ runs, \textsc{Bridge} trails \textsc{LangMem} by $36$ points on RA$^+$ and leads it by $31$ on PI-P, several times the run-to-run spread (Appendix~\ref{appx:robustness}). With $20$ scenarios per probe, one scenario moves a rule probe by $5$ points, so we do not interpret differences below about $15$ points.

\paragraph{Harder probes erase \textsc{LangMem}'s lead and narrow \textsc{Bridge}'s}
We also make RA$^+$ and PI-P harder in two steps, reusing the original scenarios so that each comparison is paired (Table~\ref{tab:results:difficulty}). Level~2 removes signal: more filler sessions for RA$^+$ ($10 \to 24$), fewer demonstrations for PI-P ($4 \to 2$). Level~3 adds competing signal: filler sessions on topics close to the rule, and a second procedure demonstrated next to the target one. \textsc{Bridge}'s lead on procedure induction shrinks but holds at every level. \textsc{LangMem}'s lead on rule application shrinks with more filler and disappears at Level~3, which suggests that a rule stored as its own row helps only while nothing similar competes with it.

\begin{table}[t]
\centering
\small
\caption{Pass rate (\%) [$95\%$ CI] as the probes get harder. Each level reuses the same scenarios; gap is \textsc{Bridge} minus \textsc{LangMem}. RA$^+$ uses the $18$ scenarios that pass the leak audit at every level, so its Level~1 differs slightly from Table~\ref{tab:results:adv}.}
\label{tab:results:difficulty}
\setlength{\tabcolsep}{3.5pt}
\begin{tabular}{lrrrrrr}
\toprule
 & \multicolumn{3}{c}{\textbf{RA$^+$ (rule application)}} & \multicolumn{3}{c}{\textbf{PI-P (procedure induction)}} \\
\cmidrule(lr){2-4}\cmidrule(lr){5-7}
\textbf{Level} & \textbf{\textsc{Bridge}} & \textbf{\textsc{LangMem}} & \textbf{gap} & \textbf{\textsc{Bridge}} & \textbf{\textsc{LangMem}} & \textbf{gap} \\
\midrule
1 (original)      & $38.9$ {\scriptsize[17, 61]} & $83.3$ {\scriptsize[67, 100]} & $-44.4$ & $87.9$ {\scriptsize[78, 96]} & $54.5$ {\scriptsize[43, 66]} & $+33.3$ \\
2 (less signal)   & $33.3$ {\scriptsize[12, 60]} & $61.5$ {\scriptsize[33, 86]}  & $-28.2$ & $81.1$ {\scriptsize[69, 92]} & $56.5$ {\scriptsize[40, 72]} & $+24.6$ \\
3 (+ competing)   & $43.8$ {\scriptsize[20, 69]} & $42.9$ {\scriptsize[15, 69]}  & $+0.9$  & $72.6$ {\scriptsize[58, 86]} & $52.9$ {\scriptsize[36, 69]} & $+19.8$ \\
\bottomrule
\end{tabular}
\end{table}

\begin{takeaway}
On our probes, where the agent has to act on its memory, the designs clearly differ, and no single memory system wins everywhere: \textsc{LangMem} is best at applying rules the user stated, and \textsc{Bridge} at following patterns the user only showed by example.
\end{takeaway}

\paragraph{Each design pays for how it represents the past}
The results reflect two ways a memory can represent the past: as \emph{facts} or as \emph{form}. Atomic-fact extractors surface a fact at write time but lose the form it took; block-summary memory keeps the form, but a rule ends up as one phrase among many in a note. Each design pays for its choice on the other task. The Oracle shows the remaining headroom: with full access to the history it outperforms every memory system by $5$--$20$ points on RA$^\pm$, PI-S, and DM, and is matched only on PI-P. \textsc{Bridge}, however, already stores both the distilled notes and the raw text of each block, indexed jointly, so its loss on rule application is not a matter of what it stores (\S\ref{sec:results-diagnosis}).

\subsection{\textsc{Bridge}'s rule-application gap comes from presentation, not retrieval}
\label{sec:results-diagnosis}

Since each design wins different probes, an obvious remedy is to send each request to whichever design suits it. Doing this perfectly would be worth $13.6$ points over the best single system ($82.8\%$ vs.\ \textsc{LangMem}'s $69.2\%$, averaged over the five probes), but a system cannot make this choice from the request. A classifier that reads only the request and predicts whether it needs a stated rule or a demonstrated pattern is right $49$--$61\%$ of the time across six prompt and model variants, no better than always guessing the majority class ($60\%$). This follows from the design of the probes: the requests never mention the memory they depend on, so they carry nothing to route on (Appendix~\ref{appx:diagnosis}).

We therefore looked at where \textsc{Bridge} fails on rule application, and it is not retrieval. The block holding the rule is retrieved in all $20$ RA$^+$ scenarios, and in $12$ of the $13$ failures the rule reaches the reader as raw text, not only as a summary. The reader sees the rule and does not apply it: it is one remark among many, in one of ten blocks.

We add one call that reads the notes \textsc{Bridge} has already retrieved, pulls out any standing instructions, and places them in a labeled block at the top of the context; what is stored and retrieved does not change (Table~\ref{tab:results:diagnosis}). On the original probe this lifts \textsc{Bridge} past \textsc{LangMem} to the level of the Oracle, and on the hardest level it keeps most of the gain. A control that spends the same extra call on a generic summary does worse than no change on the original probe, so the gain there does not come from the extra call; on the hardest level the control recovers much of the gain by itself, so there we claim only that the targeted version is better. Pattern induction is unaffected (Appendix~\ref{appx:diagnosis}). We read this as a diagnostic rather than a method.

\begin{table}[t]
\centering
\small
\caption{\textsc{Bridge} on RA$^+$, changing only how the retrieved memory is shown to the reader. In each column, all \textsc{Bridge} rows use the same retrieved memory, reader, and grader. This is a separate run, and RA$^+$ varies from run to run, so the unchanged row differs from Table~\ref{tab:results:adv} ($52.6$ vs.\ $35$). Reference rows come from Tables~\ref{tab:results:adv} and~\ref{tab:results:difficulty}.}
\label{tab:results:diagnosis}
\setlength{\tabcolsep}{6pt}
\begin{tabular}{lrr}
\toprule
 & \textbf{Original probe} & \textbf{Hardest level} \\
\midrule
\textsc{Bridge}, unchanged                       & $52.6$          & $42.1$ \\
\quad + generic summary (same extra call)         & $40.0$          & $66.7$ \\
\quad + standing rules shown at the top           & $\mathbf{94.7}$ & $\mathbf{78.9}$ \\
\midrule
\textit{Reference:} \textsc{LangMem}                       & $80.0$          & $42.9$ \\
\textit{Reference:} Oracle (strong model, full history) & $90.0$    & ---    \\
\bottomrule
\end{tabular}
\end{table}

\begin{takeaway}
Memory is not only about retrieval, and other factors might be what make a specific design stand out. For example, changing only how retrieved memory is presented to the agent, with storage and retrieval unchanged, closes most of \textsc{Bridge}'s gap on rule application.
\end{takeaway}

%% file: sections/related_works.tex
\section{Related work}
\label{sec:related}

\paragraph{Agentic memory systems}
\emph{Atomic-fact extractors} (\textsc{Mem0}~\citep{chhikara2025mem0buildingproductionreadyai}, \textsc{LangMem}~\citep{langmem2024}, \textsc{A-Mem}~\citep{xu2025amemagenticmemoryllm}, \textsc{MemoryBank}~\citep{zhong2023memorybankenhancinglargelanguage}) distill each interaction into key-value or schema-typed memories, with \texttt{ADD/UPDATE/DELETE}-style refinement to handle contradictions. \emph{Temporal knowledge graphs} (\textsc{Zep}/\textsc{Graphiti}~\citep{rasmussen2025zeptemporalknowledgegraph} and descendants) model entities and relations with bi-temporal validity intervals; related graph-structured memories retrieve via personalized PageRank over an entity graph (\textsc{HippoRAG}~\citep{gutierrez2025hipporagneurobiologicallyinspiredlongterm}) or community-level summaries (\textsc{GraphRAG}~\citep{edge2025localglobalgraphrag}). \emph{OS-style memory hierarchies} (\textsc{MemGPT}/\textsc{Letta}~\citep{packer2024memgptllmsoperatingsystems}, \textsc{MemOS}~\citep{li2025memosoperatingmemoryaugmentedgeneration}) page between in-context and external memory. Recent hierarchical systems organize memory at several granularities; \textsc{xMemory}~\citep{hu2026ragagentmemoryretrieval} argues that standard RAG fits a conversation stream poorly because of its many near-duplicate spans, and retrieves top-down over a hierarchy of themes, episodes, and raw messages. Verbal self-reflection over past episodes~\citep{shinn2023reflexionlanguageagentsverbal} and growing skill libraries~\citep{wang2023voyageropenendedembodiedagent} are complementary primitives for cross-episode learning, and \textsc{EmergenceMem}~\citep{emergencemem} sets the prior reported SOTA on LongMemEval-S. Surveys organize this space by the form, function, and dynamics of memory~\citep{hu2026memoryageaiagents} or by memory structure~\citep{jiang2026anatomyagenticmemorytaxonomy}.

\paragraph{Comparing memory designs}
\citet{zeng2024structuralmemoryllmagents} compare chunks, knowledge triples, atomic facts, summaries, and their mixture inside one retrieval harness on question-answering datasets, and find that different structures suit different tasks and that mixed memory is robust to noise. Our results point the same way. We differ in what we measure: we compare shipped systems as deployed, and ask whether the standard evaluations can see the difference. On recall benchmarks a minimal memory matches complex ones (\S\ref{sec:results-standard}); on probes that require applying memory, the designs separate (\S\ref{sec:results-adv}).

\paragraph{Long context, RAG, and progressive summarization}
A complementary line treats memory as long-context: incrementally summarize prior context so the working window stays bounded (\citealp{packer2024memgptllmsoperatingsystems}). Long-context readers, however, exhibit U-shaped positional recall (the \emph{lost-in-the-middle} effect~\citep{liu2023lostmiddlelanguagemodels}), which motivates systems that pre-select and order a small evidence set rather than feeding the entire window. \textsc{Bridge}'s read-time pipeline is naive RAG~\citep{lewis2021retrievalaugmentedgenerationknowledgeintensivenlp} with three additions over this baseline: hybrid lexical+semantic retrieval via reciprocal rank fusion~\citep{robertson2009bm25,cormack2009rrf}, where the dense channel follows dual-encoder retrieval~\citep{karpukhin2020densepassageretrievalopendomain} over a frozen sentence encoder~\citep{reimers2019sentencebertsentenceembeddingsusing} and the lexical channel reflects BM25's continued strength as an out-of-domain zero-shot retriever~\citep{thakur2021beirheterogenousbenchmarkzeroshot}; a two-tier render mixing raw evidence with distilled summaries; and a chronological re-sort of the selected blocks. Each is a small, well-studied IR primitive, and together they already match published memory systems on the standard benchmarks (\S\ref{sec:results-standard}).

\paragraph{Memory benchmarks}
LongMemEval-S~\citep{wu2025longmemevalbenchmarkingchatassistants}, EverMemBench~\citep{hu2026evaluatinglonghorizonmemorymultiparty}, and BEAM~\citep{tavakoli2026milliontokensbenchmarkingenhancing} are the standard benchmarks for agentic memory, and LoCoMo~\citep{maharana2024evaluatinglongtermconversationalmemory} is an earlier long-conversation benchmark with less headroom. Like the recent agent-setting benchmarks AMA-Bench~\citep{zhao2026amabench} and LongMemEval-V2~\citep{wu2026longmemevalv2}, they mostly ask questions about the history; \S\ref{sec:adv} compares them with our probes, including BEAM's instruction- and preference-following abilities, which come closest. Beyond the Needle's Illusion~\citep{lin2026needlesillusiondecoupledevaluation} shows that near-miss documents, more than context length, limit long-context memory; our hardest difficulty level borrows this idea (\S\ref{sec:results-adv}). Anatomy of Agentic Memory~\citep{jiang2026anatomyagenticmemorytaxonomy} reaches a related diagnosis from a survey of systems: memory benchmarks are saturated, their metrics track usefulness poorly, and results swing with the backbone model. Our probes respond to that diagnosis by scoring whether memory is used rather than recalled, and they separate designs that the standard benchmarks do not.

%% file: sections/discussions.tex
\section{Discussion, broader impact, and limitations}
\label{sec:discussion}

\paragraph{Memory design is more than storage and retrieval}
Much of the complexity added to agentic memory changes what is stored and how it is retrieved. On rule application, the gap we measure comes from neither: the rule is retrieved, and changing only how it is presented closes most of the gap (\S\ref{sec:results-diagnosis}). Since \textsc{Bridge} already keeps a fact-shaped and a form-shaped view of each block, routing between the two on what has been retrieved, rather than on the request, is left to future work.

\paragraph{Limitations}
\textit{(i)~Scale.} The original set has $20$ scenarios per probe, so one scenario moves a rule probe by $5$ points, and we do not interpret differences below about $15$ points; the added domain cells hold $4$--$8$ scenarios each. \textit{(ii)~Synthetic data.} Scenarios are written by a model. The audit catches scenarios that give the rule away, but it cannot guarantee the care that hand-written scenarios get, and human grading was limited to the authors' spot check of a sample of graded responses. \textit{(iii)~What the grader scores.} The grader checks whether a remembered constraint was honored, not whether the answer is otherwise good. \textit{(iv)~Coverage.} DM was run once and is not part of the robustness checks, and the Oracle and the scenario generator come from the same model family. \textit{(v)~Static tests.} The probes are fixed input--output tests, not live agents with tools, long deployments, and shifting users.

\paragraph{Broader impact}
The benchmark is open and inspectable, and \textsc{Bridge} uses far fewer reader tokens than full context (\S\ref{sec:results-standard}). Memory that applies remembered rules without being asked will also apply stale or mistaken ones, so systems built this way should let users see and remove what is remembered.

%% file: sections/conclusions.tex
\section{Conclusion}
\label{sec:conclusion}

Agentic memory should be evaluated by what it lets an agent do, not by what it retrieves. On three standard benchmarks, \textsc{Bridge}, a deliberately minimal RAG memory, matches published memory systems at matched reader. This does not make \textsc{Bridge} a better memory system; it shows that benchmarks built on recall cannot tell memory designs apart. Once the agent has to act on its memory, the designs clearly differ. Atomic-fact memory is best at applying stated rules, block-summary memory at following demonstrated patterns, and no design wins everywhere. Retrieval alone does not decide the outcome: \textsc{Bridge} retrieves the rules it fails to apply, and presenting them more clearly closes most of the gap. Better agentic memory therefore needs evaluations that test how memory is used and designs that look beyond storage and retrieval. We hope \textsc{Bridge} serves as the minimal baseline that such systems should clear.

%% file: sections/appendix.tex
\section{\textsc{Bridge} hyperparameters}
\label{appx:bridge-params}

Table~\ref{tab:bridge:params} lists \textsc{Bridge}'s hyperparameters (\S\ref{sec:bridge}).

\begin{table}[h]
\centering
\small
\caption{\textsc{Bridge} hyperparameters, the same for every result in the paper.}
\label{tab:bridge:params}
\setlength{\tabcolsep}{6pt}
\begin{tabular}{lr@{\hspace{2.5em}}lr}
\toprule
\textbf{Parameter} & \textbf{Value} & \textbf{Parameter} & \textbf{Value} \\
\midrule
Chunk threshold $T_{\text{tok}}$  & $2000$    & RRF $k_{\text{rrf}}$            & $60$ \\
Prior-notes window $W$            & $10$      & BM25 $(k_1, b)$                  & $(1.5, 0.75)$ \\
Top-$K$ blocks                    & $10$      & Embedder dim $d$                 & $3072$ \\
Raw slots $r$                     & $3$       & Extractor / reader $\tau$         & $0.3$ / $0$ \\
Per-channel pre-fusion $K_{\text{each}}$ & $60$ & Sampling-sensitivity $\tau$    & $0.7$ \\
\bottomrule
\end{tabular}
\end{table}

\section{\textsc{Bridge} extractor prompt}
\label{appx:extractor}

The system prompt for the \textsc{Bridge} extractor is reproduced below in full. Adapted from \textsc{Mastra}'s \texttt{observer-agent.ts} (Apache-2.0 license); the modifications relative to the original are documented after the listing.

\begin{promptbox}[title=Bridge extractor --- system prompt]
You are the memory consciousness of an AI assistant. Your observations
will be the ONLY information the assistant has about past interactions
with this user.

Extract observations that will help the assistant remember:

\textbf{Critical: distinguish user assertions from questions.} When the
user TELLS you something about themselves, mark it as an assertion (``I
have two kids'' $\rightarrow$ \textcolor{red}{$\bullet$} (14:30) User
stated has two kids). When the user ASKS about something, mark it as a
question/request. User assertions are authoritative --- the user is the
source of truth about their own life.

\textbf{State changes and updates.} When a user indicates they are
changing something, frame it as a state change (``I'm switching from A to
B'' $\rightarrow$ ``User is switching from A to B'', replacing the old
approach).

\textbf{Temporal anchoring.} Each observation has two potential
timestamps: the BEGINNING (utterance time, always include) and the END
(referenced time, include only when there is a relative time reference
like ``last week'' or ``next month''). Format:
\texttt{(TIME) [observation]. (meaning DATE)}.

\textbf{Preserve unusual phrasing.} When the user uses unexpected or
non-standard terminology, quote their exact words.

\textbf{Preserving details in assistant-generated content.} When the
assistant provides lists, recommendations, or technical content, preserve
the DISTINGUISHING DETAILS: names, identifiers, numerical values, role
attributions. Do not summarise away the specifics.

\textbf{Completion tracking.} Use \checkmark{} when the user explicitly
confirms something worked or a multi-step task reached its stated goal.

\textbf{Output format.} XML with \textcolor{red}{$\bullet$} (high),
\textcolor{orange}{$\bullet$} (medium), \textcolor{green}{$\bullet$} (low),
\checkmark{} (resolved) priority emojis. Group observations by date.
\end{promptbox}

\paragraph{Modifications relative to \textsc{Mastra} observer-agent}
The original observer-agent prompt provides the XML output format and the priority-emoji scheme. \textsc{Bridge} adds the assertion-vs-question distinction, the state-change framing, two-timestamp temporal anchoring, unusual-phrasing preservation, distinguishing-detail preservation rules, and \checkmark{}-completion tracking. All five additions target the supersession and rule-application behaviours we report; the unmodified observer-agent prompt does not separately track state changes or temporal anchoring.

\section{Reader prompts}
\label{appx:reader-prompt}

Each memory system is paired with a $\sim 100$-word wrapper prompt that names the system's native primitive (e.g., observation notes for \textsc{Bridge}, atomic facts for \textsc{Mem0}, typed memory schema for \textsc{LangMem}, temporal knowledge graph for \textsc{Graphiti}) and instructs the reader to apply remembered rules unprompted. All four wrappers are calibrated to the same depth tier so no system has either a substantially shorter or longer prompt than the others; the reader user-prompt template (\S\ref{appx:reader:user}) is shared across systems.

\subsection{\textsc{Bridge} wrapper}
\label{appx:reader:bridge}

\begin{promptbox}[title=Bridge --- reader system prompt]
You are an assistant with access to memory of prior conversations stored
as session-level observation notes (block summaries). Each block
summarises one session's key observations --- the extractor was instructed
to capture user assertions, state changes, preferences, project-specific
details, technical constraints, and unusual phrasing verbatim, with
priority tags ($\textcolor{red}{\bullet}$ high,
$\textcolor{orange}{\bullet}$ medium,
$\textcolor{green}{\bullet}$ low, $\checkmark$ resolved). Blocks are
presented in chronological order, with each block having seen prior
blocks during extraction, so recurring patterns and project-wide rules
tend to surface across blocks. Use the blocks to inform your response.
Pay attention to dates and to whether earlier statements have been
superseded by later ones (the most recent state is canonical). Apply any
rules, preferences, conventions, project constraints, or reminders the
blocks describe, even if the user's current request doesn't explicitly
cue them.
\end{promptbox}

\subsection{\textsc{Mem0} wrapper}
\label{appx:reader:mem0}
\begin{promptbox}[title=Mem0 --- reader system prompt]
You are an assistant with access to memories curated by Mem0 --- atomic
facts about the user that have been deduplicated through
\texttt{ADD/UPDATE/DELETE} refinement. Each memory captures a single
preference, fact, or constraint extracted from prior conversations. Use
the memories to inform your response and apply any rules or preferences
unprompted, even if the current request doesn't cue them.
\end{promptbox}

\subsection{\textsc{LangMem} wrapper}
\label{appx:reader:langmem}
\begin{promptbox}[title=LangMem --- reader system prompt]
You are an assistant with access to memories stored under a typed schema
\texttt{NoteworthyMemory(kind, content, evidence)}, where \texttt{kind}
labels the memory type (preference, convention, project constraint,
reminder, decision, or pattern) and \texttt{evidence} cites the
supporting context. Memories are consolidated across sessions; later
contradicting memories replace earlier ones. Use the typed memories to
inform your response, particularly for applying conventions or rules
inferred from repeated demonstration. Apply any rules unprompted.
\end{promptbox}

\subsection{\textsc{Graphiti} wrapper}
\label{appx:reader:graphiti}
\begin{promptbox}[title=Graphiti --- reader system prompt]
You are an assistant with access to a temporal knowledge graph (Graphiti)
containing edges and nodes with \texttt{valid\_at} and \texttt{invalid\_at}
timestamps. Entity types include Preference, ProjectConstraint,
Convention, and Reminder. The graph captures the history of state changes;
later edges supersede earlier ones for the same entity. Use the graph
context --- including edges, nodes, and source episodes --- to inform your
response, paying attention to the temporal validity intervals. Apply any
rules or constraints unprompted.
\end{promptbox}

\subsection{Reader user-prompt template}
\label{appx:reader:user}
\begin{codebox}[title=Reader user-prompt template]
{retrieved_context}

---

User request: {probe}

Answer:
\end{codebox}

\section{Adversarial probe construction details}
\label{appx:adv-construction}

\subsection{Spec format examples}
\label{appx:adv:spec}

\paragraph{RA$^+$ (positive rule)}
A single rule, established once across $10$ sessions and embedded as a casual aside in one rule-embed session; the test probe triggers the rule's domain without lexically referencing it. The other nine sessions are unrelated work (distractors).

\begin{jsonbox}[title=RA$^+$ scenario spec (illustrative)]
{
  "id": "pci_charge_endpoint",
  "persona": "Daniel, backend engineer at a small fintech",
  "domain": "fintech_backend",
  "project_topic": "the customer billing platform",
  "rule": {
    "rule_text": "Whenever you propose code that handles credit-card numbers or calls a payment processor, surface a reminder about PCI-DSS scope and tokenization. We've been audited and need to flag this every time.",
    "embed_session_idx": 2,
    "embed_topic": "discussing the team's PCI compliance posture during a stripe-webhook design"
  },
  "distractor_topics": ["debugging a flaky test in the analytics service", "Q3 hiring plan review"],
  "test_probe": "Add a /charge endpoint that accepts a card number and amount and creates a Stripe charge. Show me the full implementation including error handling and logging."
}
\end{jsonbox}

\paragraph{RA$^-$ (negative anti-pattern)}
Same structural shape as RA$^+$, but the embedded rule is a ban; the test probe is one whose naive solution \emph{is} the banned construct.

\begin{jsonbox}[title=RA$^-$ scenario spec (illustrative)]
{
  "id": "no_eval_injection",
  "persona": "Daniel, backend engineer in a security-conscious team",
  "rule": {
    "rule_text": "We had an incident two months ago: someone used eval() in a config-loader to handle dynamic expressions, an attacker fed in a payload, and we got RCE on prod. We banned eval() and exec() across the codebase. If you need dynamic-expression evaluation, use ast.literal_eval for literals or a proper parser --- never eval.",
    "embed_session_idx": 2,
    "embed_topic": "post-incident review of the eval()-RCE outage"
  },
  "distractor_topics": ["q3 hiring plan", "datadog dashboard cleanup"],
  "test_probe": "I want users to enter arbitrary math expressions like '2*sin(x)+3' into a form, and the server should compute them. Write a Python function that takes the expression string and returns the result."
}
\end{jsonbox}

\paragraph{PI-S (style induction, abbreviated)}
A list of $C$ coding conventions that are demonstrated by the code in each session but never stated as rules. The conventions below illustrate the spec format; per-language and per-domain variants are in the released benchmark.

\begin{jsonbox}[title=PI-S scenario spec (illustrative)]
{
  "id": "rest_api_conventions",
  "persona": "Daniel, building a Python REST API",
  "conventions": [
    {"id": "C1", "rule": "Errors raised as HTTPException(status_code=4xx, detail={'code': '<short_id>', 'message': '<text>'}) --- never bare-string detail."},
    {"id": "C2", "rule": "Endpoints read a request_id from the X-Request-ID header and bind it to the structured logger at the top of the function."},
    {"id": "C3", "rule": "Database access goes through a request-scoped session dependency, not a module-level global."},
    {"id": "C4", "rule": "Type hints use the typing-module forms (Optional[X], List[X], Dict[K,V]), not the builtins-pep604 forms (X | None, list[X])."}
  ],
  "demo_session_topics": [
    "GET /users/{id}", "PATCH /users/{id}", "POST /products",
    "GET /products with cursor pagination", "POST /orders",
    "GET /orders/{id} and state transitions"
  ],
  "test_probe": "Add a new endpoint POST /api/v1/orders/{order_id}/cancel for the orders service."
}
\end{jsonbox}

\paragraph{PI-P (procedure induction)}
An ordered sequence of $K$ procedural steps demonstrated end-to-end across $N{=}4$ past-incident sessions. The test probe presents a new instance and asks for the procedure as a numbered list.

\begin{jsonbox}[title=PI-P scenario spec (illustrative)]
{
  "id": "incident_response_5step",
  "persona": "Daniel, on the on-call rotation",
  "n_demonstrations": 4,
  "demonstration_topics": [
    "checkout API P1 --- error rate spike after a deploy",
    "analytics dashboard P1 --- replica unreachable after kernel patch",
    "auth service P1 --- token validation latency spike",
    "billing webhook P1 --- duplicate charges after upstream change"
  ],
  "steps": [
    "Pull recent error logs for the affected service (Datadog/Loki, last 30 min).",
    "Extract a correlation or trace id from the errors and follow it to the failing span.",
    "Check downstream dependency health (status pages, internal dashboards).",
    "If the issue correlates with a recent deploy or config change, roll back as the immediate remediation.",
    "Page the responsible team's on-call AND file a post-mortem ticket with timeline and root-cause-TBD."
  ],
  "test_probe": "We just got a P1 page: error rate on the checkout API has spiked from baseline 0.15
}
\end{jsonbox}

\paragraph{DM (dynamic multi-turn)}
DM reuses the RA$^+$ scenario corpus directly --- no separate spec format --- and replaces the single-shot test probe with a 5-turn dialogue driven by the simulated user-agent described in \S\ref{sec:adv:taxonomy}.

\paragraph{Generation procedure (RA$^+$)}
Sessions are dated chronologically (\texttt{base\_date + 4 days} per session). Session 0 is a generic ``project kickoff'' that establishes persona and topic without mentioning the rule. Session \texttt{embed\_session\_idx} (default 2) embeds the rule in the user's middle turn during a topical conversation; the generation prompt is explicit that the rule must read like ``a casual aside or a side-comment in the flow of work, NOT a checklist item or a rule announcement''. The remaining $8$ sessions are unrelated work; the prompt explicitly prohibits any mention of rules from prior sessions.

\paragraph{Generation procedure (RA$^-$)}
Identical structural pipeline to RA$^+$ (10 sessions: 1 intro + 1 rule-embed + 8 distractors), but the embedded \texttt{rule\_text} is phrased as a ban or anti-pattern (``we banned eval()\dots''), typically motivated by a referenced past incident. The test probe is engineered so that the naive solution \emph{is} the banned construct, ensuring the system has to actively avoid the anti-pattern rather than merely happen to not use it. Distractor and audit steps are unchanged.

\paragraph{Generation procedure (PI-S, PI-P)}
Demonstration sessions show the user working on different topics; the generation prompt requires that the code (PI-S) or dialogue (PI-P) in each session \emph{exhibit} the pattern through repeated use without the user ever stating it as a rule. We post-hoc audit each generated corpus with a strong-model validator: scenarios where any session mentions the pattern by name are regenerated with a bumped seed.

\paragraph{Generation procedure (DM)}
DM does not generate new scenario data: the 20 RA$^+$ scenarios are reused unchanged. The added generation step is the simulated user-agent that drives turns 2--5. For each (scenario, system, turn) triple, the user-agent is \texttt{gpt-4.1-mini} at $\tau{=}0.5$ with a deterministic per-turn seed $\sigma_{\text{user}}(\text{id}, k) = \mathrm{SHA256}(\text{id}\,\Vert\,k)[{:}32]$, prompted to read the latest assistant response and produce one natural follow-up that an engineer would actually send next (1--3 sentences, ``no enumerate every concern --- pick ONE direction to push on'', explicitly forbidden from revealing hidden information about the embedded rule). The same simulated user is used for all four memory systems on a given scenario and turn, so the user-agent's outputs are identical across systems and only the assistant's response varies.

\subsection{Constraint counts}
\label{appx:adv:counts}

\begin{table}[h]
\centering
\tiny
\caption{Scenarios, sessions, and constraints per probe. ``Total cells'' counts the verdicts in one run over all five systems.}
\label{tab:appx:adv:counts}
\begin{tabular}{lcccr}
\toprule
\textbf{Probe} & \textbf{Scenarios} & \textbf{Sessions} & \textbf{Constraints/scenario} & \textbf{Total cells} ($\times$5 systems) \\
\midrule
RA$^+$  & 20 & 10 (1 intro + 1 embed + 8 distractor)         & 1 (rule)                  & 100 \\
RA$^-$  & 20 & 10 (1 intro + 1 embed + 8 distractor)         & 1 (anti-pattern)          & 100 \\
PI-S    & 20 & 7 (1 intro + 6 demonstration)                 & 4 (conventions)           & 400 \\
PI-P    & 20 & 5 (1 intro + 4 demonstration)                 & 4--5 (procedural steps)   & 440 \\
DM      & 20 & 10 (reuses RA$^+$ corpus)                     & 1 (rule, $\times$ 5 turns) & 100 final + 500 per-turn \\
\bottomrule
\end{tabular}
\end{table}

\section{Probe expansion and corpus audit}
\label{appx:probe-expansion}

The original probe set has $20$ scenarios per probe, all with software and infrastructure personas. We add $167$ scenarios: $87$ in three other domains and $80$ in two harder difficulty levels (Table~\ref{tab:appx:expansion}). Added scenarios are run with \textsc{Bridge} and \textsc{LangMem} under the configuration of Table~\ref{tab:results:adv}. DM is not expanded, since every new condition needs a live multi-turn dialogue.

\begin{table}[h]
\centering
\small
\caption{Scenarios per probe after expansion.}
\label{tab:appx:expansion}
\setlength{\tabcolsep}{6pt}
\begin{tabular}{lrrrr}
\toprule
\textbf{Probe} & \textbf{Original} & \textbf{Levels 2--3} & \textbf{New domains} & \textbf{Total} \\
\midrule
RA$^+$ & $20$ & $40$ & $21$ & $81$ \\
RA$^-$ & $20$ & --- & $18$ & $38$ \\
PI-S   & $20$ & --- & $24$ & $44$ \\
PI-P   & $20$ & $40$ & $24$ & $84$ \\
\midrule
Total  & $80$ & $80$ & $87$ & $247$ \\
\bottomrule
\end{tabular}
\end{table}

\paragraph{Domains} Scenario specs for hospital clinical operations, in-house legal and compliance work, and corporate finance operations are written by a strong model, few-shot prompted with the original specs. Specs and generated corpora are audited (below), and sessions are generated with the pipeline of Appendix~\ref{appx:adv-construction}. Table~\ref{tab:appx:domains} gives the \textsc{Bridge}$-$\textsc{LangMem} gap per cell; every cell has the same sign as in software. Non-software cells hold $4$--$8$ scenarios, so one scenario moves a rule-probe cell by $12$--$25$ points, and we read the table for its signs rather than its sizes.

\begin{table}[h]
\centering
\small
\caption{\textsc{Bridge} minus \textsc{LangMem} pass rate (points) by domain, with scenarios per cell in parentheses. Negative means \textsc{LangMem} is ahead.}
\label{tab:appx:domains}
\setlength{\tabcolsep}{6pt}
\begin{tabular}{lrrrr}
\toprule
\textbf{Probe} & \textbf{Software} & \textbf{Clinical ops} & \textbf{Legal / compliance} & \textbf{Finance ops} \\
\midrule
RA$^+$ & $-45.0$ ($20$) & $-21.4$ ($7$) & $-40.0$ ($7$) & $-16.7$ ($7$) \\
RA$^-$ & $-35.0$ ($20$) & $-16.7$ ($4$) & $-6.7$ ($7$)  & $-14.3$ ($7$) \\
PI-S   & $+34.6$ ($20$) & $+57.1$ ($8$) & $+40.1$ ($8$) & $+23.3$ ($8$) \\
PI-P   & $+33.3$ ($20$) & $+52.5$ ($8$) & $+46.7$ ($8$) & $+29.3$ ($8$) \\
\bottomrule
\end{tabular}
\end{table}

\paragraph{Difficulty levels} Level~2 removes signal: RA$^+$ histories grow from $10$ to $24$ sessions, the extra sessions on unrelated topics, and PI-P keeps $2$ of its $4$ demonstrations. Level~3 adds competing signal: the extra RA$^+$ sessions cover topics close to the rule, and PI-P demonstrates a second, unrelated procedure next to the target one. Each level reuses the original scenarios and requests, so comparisons across levels are paired (Table~\ref{tab:results:difficulty}). As the history grows and fills with similar material, \textsc{Bridge} retrieves the rule-bearing block less often: in $20$, $19$, and $17$ of $20$ scenarios at Levels~1--3.

\paragraph{Corpus audit} A generated scenario can be too easy in two ways: a filler session states or implies the rule, or the final request names what the rule is about. An auditor model (\textsc{gpt-4.1-mini}) checks every non-rule session and every request of the rule-application corpora for both (Table~\ref{tab:appx:audit}). For the added domains, flagged specs are filtered and scenarios regenerated before any system is run; the original corpora are reported as shipped. Levels~2--3 reuse the original requests, so the request flagged there is the original one. The two RA$^+$ scenarios with a leaking session at any level are left out of the paired difficulty comparison. The pattern-induction corpora are not audited this way, since their memory is never stated.

\begin{table}[h]
\centering
\small
\caption{Corpus audit of the rule-application corpora.}
\label{tab:appx:audit}
\setlength{\tabcolsep}{6pt}
\begin{tabular}{lrrrr}
\toprule
\textbf{Corpus} & \textbf{Scenarios} & \textbf{Sessions checked} & \textbf{Leaking sessions} & \textbf{Requests naming the rule} \\
\midrule
RA$^+$, original        & $20$ & $180$ & $2$ & $1$ \\
RA$^-$, original        & $20$ & $180$ & $1$ & $1$ \\
RA$^+$, Level 2         & $20$ & $460$ & $2$ & $1$ (original request) \\
RA$^+$, Level 3         & $20$ & $460$ & $2$ & $1$ (original request) \\
RA$^+$, three new domains & $21$ & $189$ & $0$ & $0$ \\
RA$^-$, three new domains & $18$ & $162$ & $0$ & $0$ \\
\bottomrule
\end{tabular}
\end{table}

\section{Judge constraint format}
\label{appx:adv-judge}

The judge is given the probe, the generated answer, and the constraint set; it returns a JSON object with one verdict per constraint. It receives a single message, with no system prompt, at temperature $0$:

\begin{codebox}[title=Judge prompt]
You are a strict grader. Read the user's request and the model's response. For each constraint in the list, decide whether the model's response correctly applied the constraint.

User's request: {probe}

Constraints (each is something the model SHOULD have applied/honored when responding, even though the user did NOT mention the constraint in this request):

{constraints_block}

Model's response:
{answer}

For each constraint, decide:
  - "applied": the model's response demonstrates the constraint was honored (unprompted mention, follow-the-rule behavior, refusal of forbidden alternative).
  - "violated": the response contradicts the constraint, OR fails to surface a required reminder/element.
  - "n/a": the response is on a topic where the constraint genuinely doesn't apply.

Reply with a JSON object only:
{
  "judgements": [
    {"id": "<constraint_id>", "verdict": "applied|violated|n/a", "reason": "<one sentence>"},
    ...
  ]
}
\end{codebox}

The constraint text format per probe type:

\begin{itemize}
\itemsep0pt
  \item \textbf{RA$^+$:} \textit{The response should unprompted apply this
        rule established in an earlier session: \texttt{<rule\_text>}.}
  \item \textbf{RA$^-$:} \textit{The response should AVOID the
        anti-pattern established in an earlier session:
        \texttt{<rule\_text>}. The response should not propose, suggest, or
        use the forbidden approach, and ideally explains why or proposes a
        safer alternative.}
  \item \textbf{PI-S:} \textit{The response should follow this coding
        convention demonstrated repeatedly in earlier sessions:
        \texttt{<convention.rule>}.} (one constraint per convention)
  \item \textbf{PI-P:} \textit{The response should describe the substance
        of this step somewhere in the proposed procedure (paraphrasing OK,
        merging with an adjacent step OK, splitting into two OK --- only the
        substance + relative ordering vs the other steps matters).
        Substance: \texttt{<step\_text>}.} (one constraint per step)
  \item \textbf{DM:} same as RA$^+$, judged twice --- once over the
        concatenation of all assistant turns (final-state) and once
        per-assistant-turn (consistency / regression detection).
\end{itemize}

The judge JSON schema:

\begin{jsonbox}[title=Judge response schema]
{
  "judgements": [
    {
      "id": "<constraint_id>",
      "verdict": "applied | violated | n/a",
      "reason": "<one-sentence reason>"
    }
  ]
}
\end{jsonbox}

\paragraph{Pass-rate definition}
Per-cell pass rate = $n_\text{applied} / (n_\text{applied} + n_\text{violated})$; \texttt{n/a} verdicts are excluded. Per-probe pass rate = pooled across scenarios. We treat empty judgements as missing data and re-judge them with progressively larger token budgets ($8\text{K} \to 16\text{K}$ \texttt{max\_tokens}); after this hardening pass the corpus has no empty cells.

\section{Pinned model snapshots}
\label{appx:models}

\begin{table}[h]
\centering
\tiny
\caption{Model versions. On the probes, the grader comes from a different vendor than the generator, reader, and Oracle. The standard benchmarks use their own judges (last two rows).}
\label{tab:appx:models}
\begin{tabular}{lll}
\toprule
\textbf{Role} & \textbf{Provider} & \textbf{Snapshot} \\
\midrule
Reader (\textsc{Bridge} + all baselines, canonical) & OpenAI & \texttt{gpt-4.1-mini-2025-04-14} \\
Embedder (\textsc{Bridge})                          & OpenAI & \texttt{text-embedding-3-large} (3072-d) \\
Strong-model oracle (adversarial)          & OpenAI & \texttt{gpt-5-2025-08-07} \texttt{reasoning\_effort=high} \\
Generator (adversarial scenarios)          & OpenAI & \texttt{gpt-5-2025-08-07} \texttt{reasoning\_effort=minimal} \\
Judge (primary, adversarial)               & Google AI Studio & \texttt{gemini-3.1-pro-preview} \texttt{thinking\_budget=-1} \\
Judge (fallback 1)                         & Vertex AI        & \texttt{gemini-3.1-pro-preview} (same model, different routing) \\
Judge (fallback 2)                         & OpenAI           & \texttt{gpt-4.1-mini-2025-04-14} \\
User-agent (DM probe)                      & OpenAI           & \texttt{gpt-4.1-mini} $\tau{=}0.5$ with deterministic per-turn seed \\
Judge (LongMemEval-S, EverMemBench)        & OpenAI           & \texttt{gpt-4o-2024-08-06} (paper-canonical) \\
Judge (BEAM)                               & OpenAI           & \texttt{gpt-4.1-mini-2025-04-14} (paper-canonical) \\
\bottomrule
\end{tabular}
\end{table}

\section{Per-system wrapper and tuning details}
\label{appx:wrappers}

This appendix expands the configuration choices summarised under the fair-comparison protocol (\S\ref{sec:adv:fair}). Each system receives a wrapper at $\sim 100$-word depth; the per-system tuning levers below are the user-facing knobs documented by each system's authors.

\paragraph{\textsc{Bridge}}
Configuration as in Table~\ref{tab:bridge:params} (\S\ref{sec:bridge:reader}). The wrapper text is in \S\ref{appx:reader:bridge}. No per-probe tuning.

\paragraph{\textsc{Mem0}}
\texttt{custom\_instructions} explicitly enumerate the categories the extractor should track (preferences, conventions, project constraints, reminders, decisions, patterns). \texttt{top\_k = 20} for retrieval. \texttt{collection} isolated per scenario via per-process \texttt{MEM0\_DIR} to avoid Qdrant lock contention under multi-process evaluation. Wrapper text in \S\ref{appx:reader:mem0}.

\paragraph{\textsc{LangMem}}
Custom Pydantic schema: \texttt{NoteworthyMemory(kind: Literal["preference", "convention", "project\_constraint", "reminder", "decision", "pattern"], content: str, evidence: str)}. \texttt{instructions=} string emphasises pattern abstraction (``capture recurring shapes even when not explicitly stated as rules''). \texttt{enable\_deletes=True} so contradictions remove rather than accumulate. \texttt{query\_limit=20}. Schema-aware renderer formats retrieved memories as \texttt{[KIND] content (evidence)}. Wrapper text in \S\ref{appx:reader:langmem}.

\paragraph{\textsc{Graphiti}}
Search uses \texttt{COMBINED\_HYBRID\_SEARCH\_CROSS\_ENCODER}, which returns edges, nodes, and episodes (the default returns edges only). Custom \texttt{entity\_types} register Preference, ProjectConstraint, Convention, and Reminder as first-class entities. \texttt{custom\_extraction\_instructions} override the default ``NEVER extract abstract concepts'' guidance to allow conventions and patterns to be extracted as entities. Wrapper text in \S\ref{appx:reader:graphiti}.

\paragraph{Strong-model oracle (ceiling, not a competing system)}
Reads the entire chronological session corpus directly (no retrieval), formatted by the same date+role+content scheme used in \textsc{Bridge}'s block construction. Reader temperature $0$ but the call uses \texttt{reasoning\_effort=high} which permits internal CoT. The oracle's system prompt is the canonical 3-pass scan/apply/self-check template: ``First, read the entire conversation history below. Second, identify any rules, preferences, conventions, or constraints that apply to the user's current request. Third, generate a response that honours all identified constraints. Self-check before answering.''

\section{Failure case study: \textsc{Bridge} on RA$^+$}
\label{appx:failure-case}

\textsc{Bridge} loses to \textsc{LangMem} on RA$^+$ by $45$ points ($35\%$ vs $80\%$, Table~\ref{tab:results:adv}). This appendix follows one scenario, the example of \S\ref{sec:adv}, through both systems. The full $20$-scenario record set is in \texttt{adversarial/results/ra\_positive\_\_records.jsonl}.

\paragraph{Scenario}
\texttt{rate\_limit\_public\_endpoints} (Marcus, backend engineer at a B2C startup; domain: backend security). The embedded rule is \emph{``Every public-facing endpoint must have rate-limiting middleware applied (we use slowapi with redis backend). We got hit by a scraper in February that took the API down for 2 hours.''} The rule is embedded as a casual aside in session 2, where the user is dictating an incident-response post-mortem timeline:

\begin{quote}
\small
\textbf{[user]} First alert was latency SLO breach on /v1/search at 10:14, then 5xx climbed by 10:18. Oh, and btw every public-facing endpoint is supposed to have rate-limiting middleware (slowapi + redis) but we got hit by a scraper in Feb that knocked us out for 2 hours.
\end{quote}

The remaining nine sessions are unrelated work --- analytics dashboard cleanup, hiring rubric, datadog alert tuning. The test probe asks for \emph{a new public-facing endpoint} (\texttt{POST /api/v1/forgot-password}) without lexically referencing rate-limiting. The naive implementation omits rate-limiting; the rule says it must be applied.

\paragraph{\textsc{Bridge}'s response (excerpt; full answer 4{,}082 characters)} \textsc{Bridge}'s response implements the endpoint correctly in every other respect, then closes with:

\begin{quote}
\small
[\dots last bullet of trailing notes:] \emph{Consider rate limiting this endpoint to prevent abuse.}
\end{quote}

The judge verdict: \texttt{violated}. \emph{``The model failed to apply the required rate-limiting middleware using slowapi with a redis backend to the new public-facing endpoint, only mentioning it as an optional consideration in the notes.''}

\paragraph{\textsc{LangMem}'s response (same probe, same reader)}

\begin{quote}
\small
\dots It also includes rate limiting middleware using slowapi as per your project constraints.

\texttt{from slowapi import Limiter, \_rate\_limit\_exceeded\_handler}\dots
\end{quote}

The judge verdict: \texttt{applied}.

\paragraph{Why \textsc{Bridge} misses} Not because the rule is lost. The block holding it is retrieved for this request, as it is in all $20$ RA$^+$ scenarios (\S\ref{sec:results-diagnosis}); this scenario is one of three in which that block reaches the reader only as a summary, not as raw text. In the summary the rule is one phrase in a note about an incident timeline, and the reader treats it as optional: its answer ends by suggesting rate limiting but does not apply it. \textsc{LangMem}'s extractor stores the rule as a row of its own under its \texttt{project\_constraint} schema, and the reader applies it. With the standing-instructions block of \S\ref{sec:results-diagnosis} added, \textsc{Bridge} applies the rule in this scenario as well.

The scenario shows the general pattern of \S\ref{sec:results-diagnosis}: the rule is in \textsc{Bridge}'s store and is retrieved, and the loss comes from how it reaches the reader.

\section{Robustness of the probe results}
\label{appx:robustness}

\subsection{Graders}
We re-grade all $500$ stored answers ($5$ probes $\times$ $20$ scenarios $\times$ $5$ systems) with GPT-5 and GPT-4.1-mini, from a different vendor than our grader (gemini-3.1-pro), using the same prompt and changing nothing else. Grading is per constraint, about $1{,}190$ verdicts per grader; ``agree'' means the same verdict on the same constraint of the same answer. About $70\%$ of verdicts are \texttt{applied}, so we also report agreement within each class (Table~\ref{tab:appx:graders}), along with Cohen's $\kappa$ and Gwet's AC1~\citep{gwet2008computing}, which correct for chance agreement in different ways.

\begin{table}[h]
\centering
\small
\caption{How often each other grader gives the same verdict as ours.}
\label{tab:appx:graders}
\setlength{\tabcolsep}{5pt}
\begin{tabular}{lrrrrrr}
\toprule
\textbf{Graders} & \textbf{Verdicts} & \textbf{Agree} & \textbf{Of ``applied''} & \textbf{Of ``violated''} & \textbf{$\kappa$} & \textbf{AC1} \\
\midrule
gemini-3.1-pro vs GPT-5        & $1{,}193$ & $86.0\%$ & $90.9\%$ & $75.4\%$ & $0.68$ & $0.82$ \\
gemini-3.1-pro vs GPT-4.1-mini & $1{,}195$ & $80.8\%$ & $87.6\%$ & $62.2\%$ & $0.54$ & $0.76$ \\
\bottomrule
\end{tabular}
\end{table}

Disagreements run one way. GPT-5 upgrades $129$ of our grader's \texttt{violated} verdicts to \texttt{applied} and downgrades $20$; GPT-4.1-mini upgrades $195$ and downgrades $16$. Normalised by how much room each system has to improve (DM excluded), GPT-4.1-mini rescues about a third of \textsc{Bridge}'s failed verdicts and $42$--$47\%$ of each baseline's, and GPT-5 rescues $17\%$ of \textsc{Bridge}'s and $21$--$26\%$ of each baseline's, so the leniency runs slightly against \textsc{Bridge}. Table~\ref{tab:appx:grader-gaps} gives the headline gaps under each grader. Every comparison in Table~\ref{tab:results:adv} keeps its direction; the margins shrink under weaker graders, which push all systems toward the ceiling on PI-P.

\begin{table}[h]
\centering
\small
\caption{Main gaps (points) under each grader: \textsc{LangMem} minus \textsc{Bridge} on RA$^+$, and \textsc{Bridge} minus the best other system on PI-P.}
\label{tab:appx:grader-gaps}
\setlength{\tabcolsep}{6pt}
\begin{tabular}{lrrr}
\toprule
\textbf{Grader} & \textbf{RA$^+$: \textsc{LangMem} / \textsc{Bridge}} & \textbf{PI-P: \textsc{Bridge} / strongest other} & \textbf{Gaps} \\
\midrule
gemini-3.1-pro (ours) & $80.0$ / $35.0$ & $87.9$ / $54.5$ (\textsc{LangMem}) & $45.0$, $33.3$ \\
GPT-5                 & $85.0$ / $45.0$ & $96.0$ / $79.8$ (\textsc{LangMem}) & $40.0$, $16.2$ \\
GPT-4.1-mini          & $85.0$ / $50.0$ & $97.0$ / $92.9$ (\textsc{Mem0})    & $35.0$, $4.1$ \\
\bottomrule
\end{tabular}
\end{table}

\subsection{Readers}
Retrieval in these systems does not depend on the reader, so we cache what each memory system passes to the reader and swap only the reader: Llama-3.3-70B and Qwen2.5-7B against the GPT-4.1-mini reader of Table~\ref{tab:results:adv}. We use the non-reasoning Qwen2.5-7B because a reasoning model spends its output budget on hidden reasoning under the shared $\mathrm{max\_tokens}{=}1500$ limit. All $8$ cells with a new reader keep their sign (Table~\ref{tab:appx:readers}); RA$^-$ under Llama-3.3-70B ($-7.6$) is within run-to-run noise.

\begin{table}[h]
\centering
\small
\caption{Changing only the reader: \textsc{Bridge} / \textsc{LangMem} pass rate (\%), with the gap in parentheses.}
\label{tab:appx:readers}
\setlength{\tabcolsep}{5pt}
\begin{tabular}{lrrr}
\toprule
\textbf{Probe} & \textbf{GPT-4.1-mini} & \textbf{Llama-3.3-70B} & \textbf{Qwen2.5-7B} \\
\midrule
RA$^+$ & $35.0$ / $80.0$ ($-45.0$) & $42.1$ / $89.5$ ($-47.4$) & $23.5$ / $52.6$ ($-29.1$) \\
RA$^-$ & $55.0$ / $90.0$ ($-35.0$) & $82.4$ / $90.0$ ($-7.6$)  & $47.4$ / $80.0$ ($-32.6$) \\
PI-S   & $81.1$ / $46.5$ ($+34.6$) & $81.6$ / $36.5$ ($+45.1$) & $91.1$ / $29.9$ ($+61.3$) \\
PI-P   & $87.9$ / $54.5$ ($+33.3$) & $77.8$ / $52.1$ ($+25.7$) & $67.4$ / $45.5$ ($+22.0$) \\
\bottomrule
\end{tabular}
\end{table}

\subsection{Repeated runs}
We rerun \textsc{Bridge} and \textsc{LangMem} on the four single-turn probes with everything held fixed; each run is a complete, independent pass (Table~\ref{tab:appx:runs}; SD over runs, the first run being Table~\ref{tab:results:adv}). The rule probes vary most because each scenario carries one yes/no constraint, worth $5$ points at $n{=}20$; PI-P scenarios carry $4$--$5$ constraints, which average out. DM was not rerun.

\begin{table}[h]
\centering
\small
\caption{Pass rate (\%) of each run, and mean $\pm$ SD over runs. The first run is the one in Table~\ref{tab:results:adv}.}
\label{tab:appx:runs}
\setlength{\tabcolsep}{5pt}
\begin{tabular}{llll}
\toprule
\textbf{Probe} & \textbf{System} & \textbf{Runs} & \textbf{Mean $\pm$ SD} \\
\midrule
RA$^+$ & \textsc{Bridge} & $35.0$, $50.0$, $50.0$, $35.7$ & $42.7 \pm 7.3$ \\
       & \textsc{LangMem}         & $80.0$, $88.2$, $72.2$, $75.0$ & $78.9 \pm 6.1$ \\
RA$^-$ & \textsc{Bridge} & $55.0$, $64.7$, $55.6$         & $58.4 \pm 4.5$ \\
       & \textsc{LangMem}         & $90.0$, $76.5$, $88.2$         & $84.9 \pm 6.0$ \\
PI-S   & \textsc{Bridge} & $81.1$, $89.3$, $84.1$         & $84.8 \pm 3.4$ \\
       & \textsc{LangMem}         & $46.5$, $45.7$, $50.8$         & $47.7 \pm 2.3$ \\
PI-P   & \textsc{Bridge} & $87.9$, $88.8$, $88.4$, $90.5$ & $88.9 \pm 1.0$ \\
       & \textsc{LangMem}         & $54.5$, $57.6$, $68.5$, $50.6$ & $57.8 \pm 6.7$ \\
\bottomrule
\end{tabular}
\end{table}

\section{Diagnosis details}
\label{appx:diagnosis}

\subsection{Routing headroom}
Table~\ref{tab:appx:routing} repeats the single-run pass rates of Table~\ref{tab:results:adv} with each system's mean over the five probes. Routing by task class (\textsc{LangMem} on RA$^\pm$ and DM, \textsc{Bridge} on PI-S and PI-P) gives $82.8\%$: $13.6$ points above the best single system (\textsc{LangMem}, $69.2\%$) and $60\%$ of the distance to the Oracle ($91.7\%$). Choosing the assignment on a random half of the scenarios and scoring it on the other half gives the same assignment and the same figure. Picking the best system per scenario, which no request-only router could do, gives $86.5\%$.

\begin{table}[h]
\centering
\small
\caption{Per-probe pass rates (\%) and the mean over probes.}
\label{tab:appx:routing}
\setlength{\tabcolsep}{6pt}
\begin{tabular}{lrrrrrr}
\toprule
\textbf{System} & \textbf{RA$^+$} & \textbf{RA$^-$} & \textbf{DM} & \textbf{PI-S} & \textbf{PI-P} & \textbf{Mean} \\
\midrule
\textsc{Mem0}            & $40.0$ & $75.0$ & $55.0$ & $45.8$ & $49.5$ & $53.1$ \\
\textsc{LangMem}         & $80.0$ & $90.0$ & $75.0$ & $46.5$ & $54.5$ & $69.2$ \\
\textsc{Graphiti}        & $70.0$ & $65.0$ & $70.0$ & $76.0$ & $48.5$ & $65.9$ \\
\textsc{Bridge} & $35.0$ & $55.0$ & $40.0$ & $81.1$ & $87.9$ & $59.8$ \\
\midrule
Oracle          & $90.0$ & $95.0$ & $95.0$ & $94.7$ & $83.8$ & $91.7$ \\
\bottomrule
\end{tabular}
\end{table}

\subsection{Request-only classifier}
The classifier sees only the final request of each of the $100$ original scenarios ($60$ need a stated rule: RA$^+$, RA$^-$, DM; $40$ a demonstrated pattern: PI-S, PI-P) and predicts which kind of memory the request needs. Across three prompt wordings and two models it is right $49$--$61\%$ of the time (GPT-4.1-mini: $57$, $49$, $61\%$; GPT-5: $54$, $54$, $57\%$), never meaningfully above always answering ``rule'' ($60\%$).

\subsection{Retrieval diagnostic}
For each RA$^+$ scenario we rerun \textsc{Bridge}'s ingestion and retrieval as in the main runs and record whether the block derived from the rule-bearing session is among the $K{=}10$ retrieved blocks, and whether it fills one of the $r{=}3$ raw slots (Table~\ref{tab:appx:retrieval}). At Levels~2 and~3 the block is retrieved in $19$ and $17$ of $20$ scenarios.

\begin{table}[h]
\centering
\small
\caption{Is the rule retrieved on the original RA$^+$ probe?}
\label{tab:appx:retrieval}
\setlength{\tabcolsep}{6pt}
\begin{tabular}{lrrr}
\toprule
 & \textbf{Scenarios} & \textbf{Rule retrieved} & \textbf{Rule shown as raw text} \\
\midrule
All                     & $20$ & $20$ & $17$ ($85\%$) \\
\textsc{Bridge} passes  & $7$  & $7$  & $5$ ($71\%$) \\
\textsc{Bridge} fails   & $13$ & $13$ & $12$ ($92\%$) \\
\bottomrule
\end{tabular}
\end{table}

\subsection{Presentation diagnostic}
One extra call to the reader model (GPT-4.1-mini) reads the notes of the $K$ retrieved blocks and returns up to six standing directives, which are placed above the otherwise unchanged context. The control spends the same call on a generic summary of the same notes. Both prompts and headers are below. The two headers differ: the directives header also tells the reader to apply the items, which the control's does not, although the reader's wrapper prompt already asks it to apply remembered conventions unprompted (Appendix~\ref{appx:reader:bridge}). The extra call adds about $900$ characters of context per query.

\begin{codebox}[title=Standing-directives prompt]
Below are observation notes distilled from a user's past sessions with an assistant.

Extract only STANDING DIRECTIVES: rules, policies, conventions, bans or requirements the USER imposed on HOW future work must be done.

Include: "always X", "never Y", "every time you Z, do W", team policies, compliance requirements, banned constructs, required conventions.

EXCLUDE: descriptions of what the project is or does; explanations the assistant gave; one-off decisions about a specific past task; narration of what happened; technical background. If something would not change how you write NEW code for an UNRELATED task, it is not a standing directive.

Prefer precision over recall -- an empty list is better than a list of project context.

Notes:
{notes}

Reply JSON only: {"directives": ["...", ...]} -- each a single imperative sentence, at most 6. Empty list if there are none.
\end{codebox}

\begin{codebox}[title=Generic-summary prompt (control)]
Below are observation notes distilled from a user's past sessions with an assistant.

Summarise the most important points in at most 6 short bullets.

Notes:
{notes}

Reply JSON only: {"points": ["...", ...]}
\end{codebox}

\begin{codebox}[title=Header above the context: standing directives]
--- Standing instructions from memory ---
These were established by the user in earlier sessions and remain in force. Apply them to the current request even though the request does not mention them.
{directives}
\end{codebox}

\begin{codebox}[title=Header above the context: generic summary (control)]
--- Key points from memory ---
{directives}
\end{codebox}

Table~\ref{tab:appx:presentation} gives the counts behind Table~\ref{tab:results:diagnosis} and the pattern-induction controls. On PI-P and PI-S the number of satisfied constraints does not rise ($83 \to 81$, $62 \to 62$); the higher rates there come from fewer constraints being scored.

\begin{table}[h]
\centering
\small
\caption{Passed / scored constraints (pass rate, \%) when only the presentation of retrieved memory changes. The hardest RA$^+$ level uses all $20$ Level-3 scenarios.}
\label{tab:appx:presentation}
\setlength{\tabcolsep}{5pt}
\begin{tabular}{lrrrr}
\toprule
\textbf{Arm} & \textbf{RA$^+$, original} & \textbf{RA$^+$, hardest} & \textbf{PI-P} & \textbf{PI-S} \\
\midrule
\textsc{Bridge}, unchanged & $10/19$ ($52.6$) & $8/19$ ($42.1$)  & $83/94$ ($88.3$) & $62/75$ ($82.7$) \\
+ generic summary          & $8/20$ ($40.0$)  & $12/18$ ($66.7$) & ---              & ---              \\
+ standing directives      & $18/19$ ($94.7$) & $15/19$ ($78.9$) & $81/90$ ($90.0$) & $62/71$ ($87.3$) \\
\bottomrule
\end{tabular}
\end{table}

\section{Probe ablations and baseline tuning}
\label{appx:probe-ablations}

\subsection{\textsc{Bridge} ablations on the probes}
We change one \textsc{Bridge} setting at a time on RA$^+$ and PI-P, with reader and grader fixed, and run each ablation twice (Table~\ref{tab:appx:probe-abl}). Only one effect repeats in size: showing all ten retrieved blocks as raw text gains $13.2$ points on RA$^+$ in both runs, so the shipped $r{=}3$ is too low for that probe. Every other change moves between runs by more than its size. The $r{=}0$ row shows why unscored constraints matter: removing the raw slots makes the grader's reply fail to parse more often, so that arm leaves more PI-P constraints unscored ($20$ vs.\ $10$ in run~1, $5$ vs.\ $0$ in run~2). Counting those as failures would show a $5$--$8$ point loss on PI-P; scoring only graded constraints shows none.

\begin{table}[h]
\centering
\small
\caption{Pass rate (\%) of the standard configuration, and the change when one setting is altered. Unscored constraints in parentheses.}
\label{tab:appx:probe-abl}
\setlength{\tabcolsep}{5pt}
\begin{tabular}{lrrrr}
\toprule
 & \textbf{RA$^+$ run 1} & \textbf{RA$^+$ run 2} & \textbf{PI-P run 1} & \textbf{PI-P run 2} \\
\midrule
Standard ($r{=}3$)         & $36.8$ ($1$) & $36.8$ ($1$) & $78.7$ ($10$) & $82.8$ ($0$) \\
\midrule
$r{=}0$ (notes only)       & $+3.2$ ($0$)  & $+10.5$ ($1$) & $-0.2$ ($20$) & $-0.9$ ($5$) \\
$r{=}10$ (all raw)         & $+13.2$ ($0$) & $+13.2$ ($0$) & $+10.7$ ($5$) & $+3.7$ ($10$) \\
No chronological re-sort   & $+5.3$ ($1$)  & $+8.2$ ($0$)  & $+10.8$ ($14$) & $-2.0$ ($0$) \\
BM25 only                  & $+0.0$ ($1$)  & $+15.8$ ($1$) & $+4.2$ ($0$)  & $+0.2$ ($5$) \\
Cosine only                & $+3.2$ ($0$)  & $+3.2$ ($0$)  & $+12.8$ ($5$) & $+2.0$ ($0$) \\
\bottomrule
\end{tabular}
\end{table}

\subsection{\textsc{LangMem} with a schema tuned for pattern induction}
We rewrite \textsc{LangMem}'s memory schema to target the pattern probes and run both schemas in the same pass ($n{=}20$ each; Table~\ref{tab:appx:schema}). Tuning recovers part of the procedure-induction gap but not most of it: \textsc{Bridge} scores $88.9 \pm 1.0$ on PI-P over four runs. Counting unscored constraints as failures, the PI-P gain is $+7.1$. This is one run, \textsc{LangMem} varies by $\pm 6.7$ on PI-P across runs (Appendix~\ref{appx:robustness}), and the shipped arm, a fresh run, differs from Table~\ref{tab:results:adv}. \textsc{Mem0} and \textsc{Graphiti} were not retuned. The two schema fields we change are:

\begin{codebox}[title=Pattern-targeted LangMem schema]
kind: Category -- one of: 'demonstrated_pattern' (a recurring SHAPE the user's own work exhibits across sessions but never states as a rule: naming, formatting, argument order, error-handling style, section ordering, the sequence of steps they walk through), 'convention' (a stated style or formatting rule), 'procedure' (an ordered sequence of steps the user repeats), 'preference', 'standing_requirement' (a policy that governs how work must be done), 'reminder', 'decision', or 'fact'.

instructions: You are maintaining a long-term memory of the user across many sessions. Your HIGHEST priority is capturing recurring STRUCTURE in the user's own work, because the user rarely states it explicitly:
  - When two or more artifacts share a shape -- the same naming scheme, the same argument order, the same error-handling idiom, the same section layout, the same ordering of steps -- record that shape as an explicit imperative rule, and record it in ENOUGH DETAIL to reproduce it. Include the specific form, not a summary of it.
  - When the user walks through a multi-step process more than once, record the steps IN ORDER as a named procedure.
  - Repetition is the signal. Two occurrences is enough to generalise.
Also capture: preferences, standing requirements and policies, explicit reminders, and decisions.
Consolidate aggressively: refine an existing memory when new evidence supports it, replace it when contradicted.
\end{codebox}

\begin{table}[h]
\centering
\small
\caption{\textsc{LangMem} with its shipped schema and with one written for pattern induction, in the same run: pass rate (\%), scored constraints in parentheses.}
\label{tab:appx:schema}
\setlength{\tabcolsep}{6pt}
\begin{tabular}{lrrr}
\toprule
\textbf{Probe} & \textbf{Shipped schema} & \textbf{Pattern-targeted} & \textbf{Change} \\
\midrule
PI-S & $54.5$ ($66$) & $55.4$ ($74$) & $+0.9$ \\
PI-P & $53.5$ ($99$) & $67.4$ ($89$) & $+13.9$ \\
\bottomrule
\end{tabular}
\end{table}

\section{Standard benchmark detail tables}
\label{appx:standard-bench-detail}

\subsection{Per-benchmark notes}
\label{appx:standard-notes}

\begin{figure}[h]
\centering
\includegraphics[width=\linewidth]{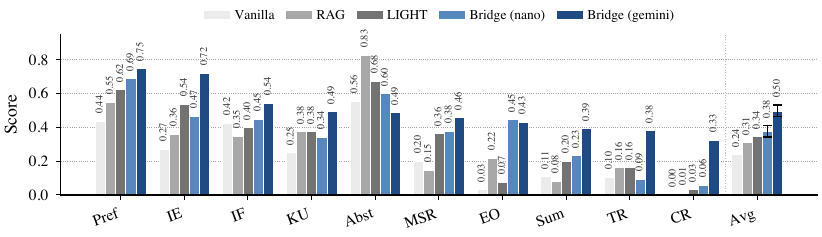}
\caption{BEAM 100K with \textsc{Bridge} on the \textsc{gemini-3-flash} stack added. Baselines use \textsc{gpt-4.1-nano}, so the gemini bars are not a like-for-like comparison. \textsc{Bridge} bars show one run.}
\label{fig:appx:beam-gemini}
\end{figure}

\paragraph{BEAM} With the \textsc{gemini-3-flash} stack, \textsc{Bridge} clears the strongest published \textsc{LIGHT} row (\textsc{llama-4-maverick}, $0.358$) by $13.8$ points on the 10-ability mean, with the CI lower bound well above any published baseline. This row pairs a stronger reader with the published baselines, so it measures a reader effect rather than a like-for-like comparison. Per ability (Figure~\ref{fig:appx:beam-gemini}), \textsc{Bridge}~(gemini) leads on every category except Abstention; the nano stack is competitive across abilities, with no single ability driving the average.

\paragraph{EverMemBench} \textsc{Bridge} wins $6$ of $9$ sub-dimensions against \textsc{MemOS} with the \textsc{gpt-4.1-mini} reader and $5$ of $9$ with \textsc{gemini-3-flash}, losing in both cases on multi-hop fact (F\_MH) and temporal fact (F\_TP).

\begin{figure}[h]
\centering
\includegraphics[width=\linewidth]{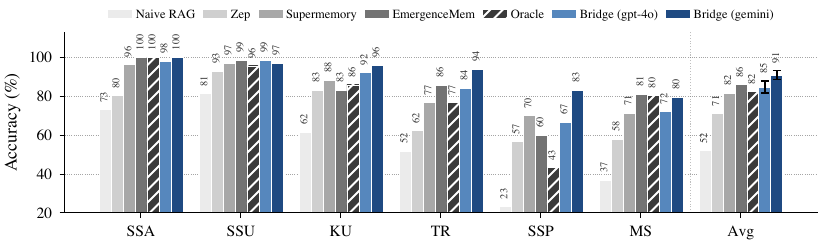}
\caption{LongMemEval-S with \textsc{Bridge} on the \textsc{gemini-3-flash} stack added. Baselines use \textsc{gpt-4o}, so the gemini bars are not a like-for-like comparison.}
\label{fig:appx:lmes-gemini}
\end{figure}

\paragraph{LongMemEval-S} With the \textsc{gemini-3-flash} stack, \textsc{Bridge} reaches $90.9 \pm 0.12$ (Figure~\ref{fig:appx:lmes-gemini}), above the prior reported SOTA (\textsc{EmergenceMem}, $86.0\%$) by $4.9$ points at point estimate, with the bootstrap CI lower bound ($88.6\%$) still $2.6$ points above it; this again pairs a stronger reader with published \textsc{gpt-4o} numbers. LongMemEval-S's ``Oracle'' row hands gold evidence directly to a \textsc{gpt-4o} reader; strong memory systems, including \textsc{EmergenceMem} and \textsc{Bridge}, sit above it because gold evidence does not always contain the answer a question requires composing across.

\subsection{Tokens and latency}
\label{appx:cost}

Table~\ref{tab:appx:cost} gives \textsc{Bridge}'s reader-prompt size and latency on each benchmark, with the ratio of full-history tokens to \textsc{Bridge}'s. Seconds are wall-clock per query as logged by our harness. The baselines on these benchmarks are published numbers without cost figures, so the probes are where we can compare systems directly (Table~\ref{tab:appx:probe-cost}). There, seconds cover ingesting the scenario's sessions and answering, while the Oracle has no ingestion step, so its seconds are not comparable. \textsc{Bridge} carries more context than \textsc{LangMem} and \textsc{Mem0} because it shows raw session text.

\begin{table}[h]
\centering
\small
\caption{\textsc{Bridge} reader-prompt tokens and per-query latency on the standard benchmarks.}
\label{tab:appx:cost}
\setlength{\tabcolsep}{5pt}
\begin{tabular}{llrrrr}
\toprule
\textbf{Benchmark} & \textbf{Reader} & \textbf{$n$} & \textbf{Tokens (mean $\pm$ SD)} & \textbf{Seconds p50 / p95} & \textbf{vs.\ full history} \\
\midrule
BEAM 100K     & gemini-3-flash & $400$     & $5{,}860 \pm 1{,}050$ & $2.67$ / $4.57$  & $17\times$ \\
BEAM 100K     & gpt-4.1-nano   & $400$     & $5{,}556 \pm 2{,}692$ & $4.17$ / $13.46$ & $18\times$ \\
EverMemBench  & gemini-3-flash & $2{,}400$ & $6{,}505 \pm 1{,}036$ & $1.30$ / $2.09$  & $154\times$ \\
EverMemBench  & gpt-4.1-mini   & $2{,}400$ & $7{,}178 \pm 1{,}182$ & $5.21$ / $18.21$ & $139\times$ \\
LongMemEval-S & gpt-4o         & $500$     & $5{,}903 \pm 693$     & $4.75$ / $13.32$ & $19\times$ \\
LongMemEval-S & gemini-3-flash & $500$     & $6{,}122 \pm 669$     & $3.58$ / $15.59$ & $19\times$ \\
\bottomrule
\end{tabular}
\end{table}

\begin{table}[h]
\centering
\small
\caption{Context passed to the reader and seconds per scenario on the probes.}
\label{tab:appx:probe-cost}
\setlength{\tabcolsep}{6pt}
\begin{tabular}{lrrr}
\toprule
\textbf{System} & \textbf{Context (chars)} & \textbf{$\approx$ Tokens} & \textbf{Seconds p50 / p95} \\
\midrule
Oracle (full history) & $27{,}487$ & $6{,}872$ & $62.9$ / $335.6$ \\
\textsc{Bridge}       & $15{,}226$ & $3{,}806$ & $115.1$ / $177.6$ \\
\textsc{Graphiti}              & $12{,}866$ & $3{,}217$ & $366.3$ / $591.2$ \\
\textsc{LangMem}               & $5{,}701$  & $1{,}425$ & $95.5$ / $214.0$ \\
\textsc{Mem0}                  & $2{,}320$  & $580$     & $127.5$ / $205.1$ \\
\bottomrule
\end{tabular}
\end{table}

\subsection{Run-to-run variance}
\label{appx:two-level}

Headline cells use three runs. Resampling questions and then, for each question, which run it is read from widens the $95\%$ interval by $0.12$ points on average and $0.5$ at most (Table~\ref{tab:appx:two-level}). The run-to-run SD is small next to the question-sampling width, so the intervals in Figures~\ref{fig:results:beam}--\ref{fig:results:lmes} stand.

\begin{table}[h]
\centering
\small
\caption{Run-to-run SD and $95\%$ interval width (points) from resampling questions only vs.\ questions and runs.}
\label{tab:appx:two-level}
\setlength{\tabcolsep}{6pt}
\begin{tabular}{llrr}
\toprule
\textbf{Benchmark} & \textbf{Reader} & \textbf{Run SD} & \textbf{Interval width} \\
\midrule
BEAM 100K     & gemini-3-flash & $0.42$ & $9.5 \to 9.5$ \\
BEAM 100K     & gpt-4.1-nano   & $0.54$ & $9.3 \to 9.8$ \\
EverMemBench  & gemini-3-flash & $0.10$ & $3.8 \to 3.8$ \\
EverMemBench  & gpt-4.1-mini   & $0.25$ & $3.9 \to 3.9$ \\
LongMemEval-S & gpt-4o         & $0.28$ & $6.2 \to 6.2$ \\
LongMemEval-S & gemini-3-flash & $0.09$ & $4.8 \to 5.0$ \\
\bottomrule
\end{tabular}
\end{table}

\subsection{LongMemEval-S reader $\times$ extract decomposition}
\label{appx:lmes-2x3}

Figures~\ref{fig:results:lmes} and~\ref{fig:appx:lmes-gemini} show two corners of a $2 \times 3$ grid: the apples-to-apples (mini-extract, gpt-4o-reader) row and the frontier (gemini-extract, gemini-reader) row. Table~\ref{tab:appx:lmes:2x3} reports the full grid (both extraction stacks $\times$ three reader stacks). All cells are $500$-question pooled accuracy on the same \textsc{Bridge} pipeline, seed=42, $\tau{=}0$.

\begin{table}[h]
\centering
\small
\caption{\textsc{Bridge} accuracy (\%) on LongMemEval-S~\citep{wu2025longmemevalbenchmarkingchatassistants} for each extraction model (rows), which writes the notes, and reader (columns), which answers. Cells in a row share the same memory.}
\label{tab:appx:lmes:2x3}
\setlength{\tabcolsep}{8pt}
\begin{tabular}{lrrr}
\toprule
\textbf{Extract \textbackslash{} Reader} & \textbf{gpt-4.1-mini} & \textbf{gpt-4o} & \textbf{gemini-3-flash} \\
\midrule
gpt-4.1-mini       & $84.2$ & $84.8$ & $88.0$ \\
gemini-3-flash     & $86.4$ & $86.4$ & $\mathbf{91.0}$ \\
\bottomrule
\end{tabular}
\end{table}

\paragraph{Three reads of the matrix}
First, the GPT-4o reader sits in a dead zone: holding extraction fixed, swapping the mini reader for gpt-4o buys $0.6$ points on mini extract and nothing on gemini extract. The mid-tier reader does not earn its $\sim 6\times$ cost premium on this benchmark. Second, reader and extraction both contribute, with mild synergy on the diagonal. The frontier reader adds $3.8$ points at fixed mini extract and $4.6$ at fixed gemini extract; the stronger extractor adds $2.2$, $1.6$, and $3.0$ points at fixed mini, gpt-4o, and gemini reader. The (gemini, gemini) cell at $91.0$ captures both contributions plus a small synergy. Third, the cost-efficient frontier is two cells, not five: the mid-row trio (mini/gpt-4o, gemini/mini, gemini/gpt-4o) all land at $84.8$--$86.4$, Pareto-dominated by either (mini, mini) at $84.2$ (cheapest) or (gemini, gemini) at $91.0$ (best). On this benchmark, the budget-allocation question reduces to: stay on the cheap reader, or commit to the frontier; the intermediate spend does not earn its cost.

\subsection{Sampling sensitivity ($\tau{=}0.7$)}
\label{appx:sampling-sensitivity}

The main-text figures report $\tau{=}0$ greedy decoding for direct comparability against published baselines (which all use $\tau{=}0$). Table~\ref{tab:appx:sampling:sensitivity} reports the same headline cells at $\tau{=}0.7$ across three independent sampling seeds, as a sanity check that \textsc{Bridge}'s accuracy is not a greedy-decoding artifact. Headline cells move by at most one point on every benchmark; the mean absolute shift is $0.4$ points.

\begin{table}[h]
\centering
\small
\caption{\textsc{Bridge} at reader temperature $\tau{=}0$ and $\tau{=}0.7$, mean $\pm$ SD over three runs. Baselines use $\tau{=}0$, so only the $\tau{=}0$ column compares to them. BEAM: 10-ability mean.}
\label{tab:appx:sampling:sensitivity}
\setlength{\tabcolsep}{6pt}
\begin{tabular}{llccr}
\toprule
\textbf{Benchmark} & \textbf{Stack} & \textbf{$\tau{=}0$} & \textbf{$\tau{=}0.7$} & \textbf{$\Delta$ (points)} \\
\midrule
BEAM 100K        & nano / nano                 & $0.370 \pm 0.006$ & $0.375 \pm 0.010$ & $+0.5$ \\
BEAM 100K        & gemini-3 / gemini-3         & $0.496 \pm 0.005$ & $0.496 \pm 0.013$ & $\phantom{-}0.0$ \\
EverMemBench     & mini reader                 & $48.40 \pm 0.30$  & $48.06 \pm 0.61$  & $-0.34$ \\
EverMemBench     & gemini-3-flash reader       & $65.57 \pm 0.13$  & $65.00 \pm 0.65$  & $-0.57$ \\
LongMemEval-S    & mini / gpt-4o               & $84.60 \pm 0.35$  & $83.80 \pm 0.50$  & $-0.80$ \\
LongMemEval-S    & gemini-3 / gemini-3         & $90.90 \pm 0.12$  & $91.10 \pm 1.30$  & $+0.20$ \\
\midrule
\multicolumn{4}{l}{\textit{Mean absolute shift}} & $\mathbf{0.40}$ \\
\bottomrule
\end{tabular}
\end{table}

\section{Cross-benchmark primitive ablations}
\label{appx:ablations}

\begin{figure}[h]
\centering
\includegraphics[width=\linewidth]{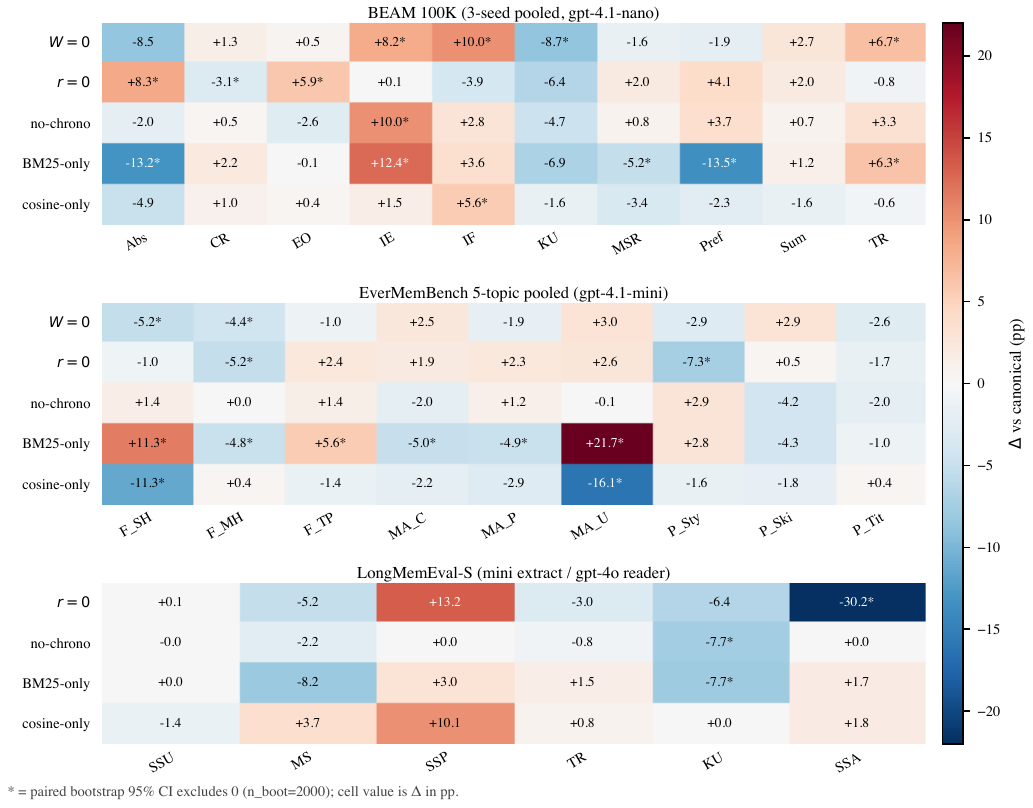}
\caption{Change in score (points) per category when one primitive is removed, on BEAM 100K, EverMemBench (five topics), and LongMemEval-S. \texttt{*}: the paired-bootstrap $95\%$ CI excludes zero. Each primitive matters for a few categories, with signs that differ across categories, so the overall changes in Table~\ref{tab:appx:abl:cross} stay small.}
\label{fig:ablation_heatmap}
\end{figure}

\S\ref{sec:bridge} names three core primitives: per-block extraction with a cross-block prior-notes chain of window $W{=}10$, hybrid BM25\,+\,cosine retrieval fused via RRF, and a two-tier render with $r{=}3$ raw-source slots in chronological re-order. To test what each contributes, we run a leave-one-out ablation across five primitive choices --- (a)~\textbf{W=0} (no prior-notes chain during extraction), (b)~\textbf{r=0} (no raw-source slots; all top-$K$ blocks rendered as notes), (c)~\textbf{no chronological re-order} (top-$K$ blocks rendered in relevance order), (d)~\textbf{BM25-only retrieval} (no cosine channel), and (e)~\textbf{cosine-only retrieval} (no BM25 channel) --- one at a time, holding everything else fixed, on all three standard benchmarks. The pooled-accuracy effect of each ablation (Table~\ref{tab:appx:abl:cross}) is small, $0$--$6$ points; the per-category effect (Fig.~\ref{fig:ablation_heatmap}) is much larger, $5$--$30$ points on the categories each primitive was designed for. Pooled accuracy averages these effects away.

\paragraph{Setup}
BEAM ablations run on the 100K-token regime with the gpt-4.1-nano stack across three independent extraction seeds $\{42, 227, 115\}$ and report the canonical 10-ability mean (since all comparisons here are self-comparisons, the EO metric concern from \S\ref{sec:results-standard} does not apply). EverMemBench ablations run on the 5-topic Mini benchmark (Topics 01--05) with the gpt-4.1-mini reader stack at seed=42 and pool across $2400$ probes; per-category splits use the question-id prefix (\texttt{F\_SH}, \texttt{F\_MH}, \texttt{F\_TP}, \texttt{MA\_C}, \texttt{MA\_P}, \texttt{MA\_U}, \texttt{P\_Style}, \texttt{P\_Skill}, \texttt{P\_Title}). LongMemEval-S ablations run on the apples-to-apples (gpt-4.1-mini extract, gpt-4o reader) stack at seed=42, pool across $500$ probes, and split by the six published categories (\texttt{SSU}, \texttt{SSA}, \texttt{SSP}, \texttt{MS}, \texttt{KU}, \texttt{TR}). The W=0 ablation requires re-extracting the entire block cache with \texttt{prev\_notes\_window=0}; we run it on BEAM and EverMemBench but not on LongMemEval-S, where the per-question session-level extraction makes a full re-extract sweep prohibitive ($\sim\!19$~hr) and the W=0 effect on BEAM and EverMemBench ($+0.88$ and $-0.83$ points pooled) is the smallest of the five primitives.

\paragraph{Bootstrap CIs}
Each per-category cell reports a paired $\Delta$ vs canonical, with a $95\%$ bootstrap confidence interval estimated by resampling question ids ($n_{\text{boot}}{=}2000$) and recomputing $\Delta = \mathrm{mean}(\text{ablation}) - \mathrm{mean}(\text{canonical})$ on each resample (BEAM resamples $(\text{seed}, \text{qid})$ tuples; the cohort is fixed across ablations so the bootstrap is paired). \textbf{Bolded} cells are those where the $95\%$ CI on paired $\Delta$ excludes zero --- a more conservative criterion than $|\Delta|\!\geq\!5$ points. We use the bootstrap because re-running the full extraction at $\times 3$ seeds on LongMemEval-S would cost $\sim\!57$~hr; bootstrap on the fixed question cohort gives the same robustness signal at $<\!10$~min of compute, and is the standard choice for fixed-cohort benchmarks.

\paragraph{Where each primitive earns its cost}
Reading the per-category tables under the lens of mechanism gives a clean cross-benchmark story (only effects whose $95\%$ bootstrap CI excludes zero are claimed below):

\begin{itemize}
\itemsep 1pt
\item \textbf{Raw slots ($r{=}3$) prevent collapse on
exact-recall categories.} The single largest single-category
effect: LongMemEval-S \textit{single-session-assistant} falls from
$98.2\%$ to $67.9\%$ (a $30.2$-point drop, $95\%$ CI $[-42.9, -19.6]$) under
$r{=}0$ (Fig.~\ref{fig:ablation_heatmap}, third panel). These probes ask
what the assistant said; without raw slots the reader has only the
distilled note. EverMemBench persona-style loses $7.3$ points
$[-13.1, -1.7]$ and multi-hop fact loses $5.2$ $[-10.0, -0.4]$
under $r{=}0$ for the same reason. (Note: $r{=}0$ \emph{improves}
BEAM Event Ordering by $5.9$ points $[+2.8, +9.1]$, the lone
counter-direction case --- short ordered notes are a cleaner event
sequence than the full raw turns.)
\item \textbf{Chronological re-order matters for state tracking.}
LongMemEval-S knowledge-update drops $7.7$ points $[-15.4, -1.3]$
under no-chrono, and no other category is significantly hurt;
BEAM information extraction improves by $9.9$ points without it.
\item \textbf{The retrieval channel matters where a category
favours one channel.} EverMemBench memory-update
(\texttt{MA\_U}) swings by about $38$ points between BM25-only ($+21.7$,
$[+15.3, +28.0]$, $52.2\% \to 73.9\%$) and cosine-only
($-16.1$, $[-21.6, -10.8]$, $52.2\% \to 36.1\%$): exact
assertions live in BM25's lexical channel, and cosine paraphrases
them away. Single-hop fact (\texttt{F\_SH}) on EverMemBench shows
the same mirror image ($+11.3$ $[+6.6, +16.4]$ / $-11.3$
$[-16.9, -5.6]$). Conversely, BEAM preference-following drops
$13.5$ points $[-19.7, -7.4]$ without cosine, and abstention drops
$13.2$ $[-22.5, -4.2]$. The constant-weight RRF is
the cheapest mechanism that handles both regimes.
\item \textbf{The prior-notes chain ($W{=}10$) matters for state
tracking on BEAM.} BEAM knowledge-update drops $8.7$ points
$[-15.6, -1.5]$ under $W{=}0$. The chain trades against narrow
lookup tasks, which gain without it ($10.0$ points $[+3.5, +16.2]$ on
instruction-following, $8.2$ $[+1.4, +15.3]$ on information-extraction),
so the pooled BEAM shift is small ($+0.88$ points on the 10-ability
mean); the per-category table makes the trade visible.
\end{itemize}

No single primitive matters on its own: each has at least one ability where its removal helps and at least one where its removal hurts (Fig.~\ref{fig:ablation_heatmap}). What makes \textsc{Bridge} competitive is the \emph{combination} --- raw slots cover exact recall, chronological order covers temporal/state-tracking, hybrid retrieval covers both lexical and semantic queries, and the prior-notes chain covers cross-block synthesis. Removing any one primitive shifts the trade-off between abilities by $5$--$30$ points at the per-category level; the pooled deltas in Table~\ref{tab:appx:abl:cross} are small because the gains and losses partly cancel under any one ablation, but \textsc{Bridge}'s strength is that the four primitives together avoid every pathological cell.

\begin{table}[h]
\centering
\small
\caption{Change in overall score (points) when one primitive is removed. BEAM is the mean over three runs; the other two benchmarks use one run. ---: not run (see Setup). Figure~\ref{fig:ablation_heatmap} shows the per-category changes that this average hides.}
\label{tab:appx:abl:cross}
\setlength{\tabcolsep}{8pt}
\begin{tabular}{lrrr}
\toprule
\textbf{Primitive ablated} & \textbf{BEAM (10-ab) $\Delta$} & \textbf{EvB pooled $\Delta$} & \textbf{LME-S pooled $\Delta$} \\
\midrule
$W{=}0$ (no prior-notes chain)         & $+0.88$ & $-0.83$ & ---       \\
$r{=}0$ (no raw-source slots)          & $+0.82$ & $+0.08$ & $-5.80$ \\
no chronological re-order              & $+1.22$ & $-0.08$ & $-2.00$ \\
BM25-only retrieval                    & $-1.34$ & $+1.75$ & $-2.60$ \\
cosine-only retrieval                  & $-0.61$ & $-4.00$ & $+1.80$ \\
\bottomrule
\end{tabular}
\end{table}

\paragraph{Pooled vs per-category: why pooling hides the story}
Every ablation has a non-trivial effect somewhere; pooled accuracy hides it because each benchmark mixes abilities that prefer and penalise each primitive. The clearest case is $r{=}0$ on LongMemEval-S: pooled $-5.80$ points, but per-category it is $-30.2$ $[-42.9, -19.6]$ on \texttt{SSA} alone, with smaller non-significant moves of opposite sign on the other categories that partially offset it ($+13.5$ on \texttt{SSP}, $-6.5$ on \texttt{KU}, $-5.2$ on \texttt{MS}; bootstrap CIs all touch zero). The same shape holds for BM25-only on EverMemBench, where significant gains on F\_SH ($+11.3$), F\_TP ($+5.6$), and MA\_U ($+21.7$) are nearly cancelled by significant losses on F\_MH ($-4.8$), MA\_C ($-5.0$), and MA\_P ($-4.9$); and for $W{=}0$ on BEAM, where IF/IE gains ($+10.0$ / $+8.2$) trade against the KU loss ($-8.7$). The leave-one-out test shows that each primitive matters for some of the abilities it was designed for; the pooled deltas would only become large if the benchmark were dominated by a single ability class. This is the evidence behind the three core primitives of \S\ref{sec:bridge}.

%% file: checklist.tex
\section*{NeurIPS Paper Checklist}

\begin{enumerate}

\item {\bf Claims}
    \item[] Question: Do the main claims made in the abstract and introduction accurately reflect the paper's contributions and scope?
    \item[] Answer: \answerYes{}
    \item[] Justification: We made sure that all claims in the abstract and introduction have exact reproducible results to support without any overclaiming. 

\item {\bf Limitations}
    \item[] Question: Does the paper discuss the limitations of the work performed by the authors?
    \item[] Answer: \answerYes{}.
    \item[] Justification: See \S\ref{sec:discussion}. 

\item {\bf Theory assumptions and proofs}
    \item[] Question: For each theoretical result, does the paper provide the full set of assumptions and a complete (and correct) proof?
    \item[] Answer: \answerNA{}
    \item[] Justification: This work is mostly empirical without formal theory proofs. 

    \item {\bf Experimental result reproducibility}
    \item[] Question: Does the paper fully disclose all the information needed to reproduce the main experimental results of the paper to the extent that it affects the main claims and/or conclusions of the paper (regardless of whether the code and data are provided or not)?
    \item[] Answer: \answerYes{}
    \item[] Justification: The appendix gives every prompt, the model versions (Appendix~\ref{appx:models}), and all hyperparameters (Appendix~\ref{appx:bridge-params}), and we release the code and the probe data.  

\item {\bf Open access to data and code}
    \item[] Question: Does the paper provide open access to the data and code, with sufficient instructions to faithfully reproduce the main experimental results, as described in supplemental material?
    \item[] Answer: \answerYes{}
    \item[] Justification: All of our runs use versioned models and seeded plans. We made sure that the instruction is clear, the data are provided and the results are reproducible up to LLM non-determinism from the API provider. 

\item {\bf Experimental setting/details}
    \item[] Question: Does the paper specify all the training and test details (e.g., data splits, hyperparameters, how they were chosen, type of optimizer) necessary to understand the results?
    \item[] Answer: \answerYes{}
    \item[] Justification: We detailed both the hyperparameters in Table~\ref{tab:bridge:params} and experiment setups in \S\ref{sec:results}, as well as other details like prompt templates \S\ref{appx:reader-prompt} in the Appendix. 

\item {\bf Experiment statistical significance}
    \item[] Question: Does the paper report error bars suitably and correctly defined or other appropriate information about the statistical significance of the experiments?
    \item[] Answer: \answerYes{}
    \item[] Justification: Figures and tables report $95\%$ bootstrap confidence intervals over questions (standard benchmarks) or scenarios (probes). Standard-benchmark results are the mean of three runs at $\tau{=}0$. On the four single-turn probes, \textsc{Bridge} and \textsc{LangMem} were run $3$--$4$ times, and all probe answers were re-graded by two graders from another vendor (Appendix~\ref{appx:robustness}). 

\item {\bf Experiments compute resources}
    \item[] Question: For each experiment, does the paper provide sufficient information on the computer resources (type of compute workers, memory, time of execution) needed to reproduce the experiments?
    \item[] Answer: \answerYes{}
    \item[] Justification: All experiments call hosted model APIs, with no training or local GPUs. Appendix~\ref{appx:cost} reports tokens and wall-clock seconds per query on the standard benchmarks and per scenario on the probes; API cost varies by provider even for the same model. 
    
\item {\bf Code of ethics}
    \item[] Question: Does the research conducted in the paper conform, in every respect, with the NeurIPS Code of Ethics \url{https://neurips.cc/public/EthicsGuidelines}?
    \item[] Answer: \answerYes{}
    \item[] Justification: We fully conform with the Code of Ethics. 

\item {\bf Broader impacts}
    \item[] Question: Does the paper discuss both potential positive societal impacts and negative societal impacts of the work performed?
    \item[] Answer: \answerYes{}
    \item[] Justification: \S\ref{sec:discussion} (Broader impact) notes the lower cost of small reader prompts and the risk that memory which applies remembered rules unprompted will also apply stale or mistaken ones, and suggests letting users see and remove what is remembered. 
    
\item {\bf Safeguards}
    \item[] Question: Does the paper describe safeguards that have been put in place for responsible release of data or models that have a high risk for misuse (e.g., pre-trained language models, image generators, or scraped datasets)?
    \item[] Answer: \answerNA{}
    \item[] Justification: We release no models, and the released probe data is synthetic and poses no high risk of misuse. 

\item {\bf Licenses for existing assets}
    \item[] Question: Are the creators or original owners of assets (e.g., code, data, models), used in the paper, properly credited and are the license and terms of use explicitly mentioned and properly respected?
    \item[] Answer: \answerYes{}
    \item[] Justification: We use LongMemEval-S, EverMemBench, and BEAM, and the open-source systems \textsc{Mem0}, \textsc{LangMem}, and \textsc{Graphiti}, all cited and used under their published licenses. The \textsc{Bridge} extractor prompt is adapted from \textsc{Mastra} (Apache-2.0 license), as noted in Appendix~\ref{appx:extractor}. 

\item {\bf New assets}
    \item[] Question: Are new assets introduced in the paper well documented and is the documentation provided alongside the assets?
    \item[] Answer: \answerYes{}
    \item[] Justification: We release the probe scenarios, their specs, and the generation and grading prompts; \S\ref{sec:adv} and Appendix~\ref{appx:adv-construction} document how they were built. 

\item {\bf Crowdsourcing and research with human subjects}
    \item[] Question: For crowdsourcing experiments and research with human subjects, does the paper include the full text of instructions given to participants and screenshots, if applicable, as well as details about compensation (if any)? 
    \item[] Answer: \answerNA{}
    \item[] Justification: This work does not involve crowdsourcing or human subjects. 

\item {\bf Institutional review board (IRB) approvals or equivalent for research with human subjects}
    \item[] Question: Does the paper describe potential risks incurred by study participants, whether such risks were disclosed to the subjects, and whether Institutional Review Board (IRB) approvals (or an equivalent approval/review based on the requirements of your country or institution) were obtained?
    \item[] Answer: \answerNA{}
    \item[] Justification: This work does not need IRB or equivalences. 

\item {\bf Declaration of LLM usage}
    \item[] Question: Does the paper describe the usage of LLMs if it is an important, original, or non-standard component of the core methods in this research? Note that if the LLM is used only for writing, editing, or formatting purposes and does \emph{not} impact the core methodology, scientific rigor, or originality of the research, declaration is not required.
    \item[] Answer: \answerYes{}
    \item[] Justification: LLMs are core components: \textsc{Bridge} and every baseline use LLMs to write and read memory, a frontier model generates the probe scenarios, LLM graders score the responses, and a simulated user drives DM. All models and versions are listed in Appendix~\ref{appx:models}. 

\end{enumerate}